# Adaptive Approach to Enhance Machine Learning Scheduling Algorithms During Runtime Using Reinforcement Learning in Metascheduling Applications

Samer Alshaer, Ala' Khalifeh, and Roman Obermaisser
*Department of Embedded Systems*
*University of Siegen*
Siegen, Germany

***Abstract*—Metascheduling in time-triggered architectures has been crucial in adapting to dynamic and unpredictable environments, ensuring the reliability and efficiency of task execution. However, traditional approaches face significant challenges when training Artificial Intelligence (AI) scheduling inferences offline, particularly due to the complexities involved in constructing a comprehensive Multi-Schedule Graph (MSG) that accounts for all possible scenarios. The process of generating an MSG that captures the vast probability space, especially when considering context events like hardware failures, slack variations, or mode changes, is resource-intensive and often infeasible. To address these challenges, we propose an adaptive online learning unit integrated within the metascheduler to enhance performance in real-time. The primary motivation for developing this unit stems from the limitations of offline training, where the MSG created is inherently a subset of the complete space, focusing only on the most probable and critical context events. In the online mode, Reinforcement Learning (RL) plays a pivotal role by continuously exploring and discovering new scheduling solutions, thus expanding the MSG and enhancing system performance over time. This dynamic adaptation allows the system to handle unexpected events and complex scheduling scenarios more effectively. Several RL models were implemented within the online learning unit, each designed to address specific challenges in scheduling. These models not only facilitate the discovery of new solutions but also optimize existing schedulers, particularly when stricter deadlines or new performance criteria are introduced. By continuously refining the AI inferences through real-time training, the system remains flexible and capable of meeting evolving demands, thus ensuring robustness and efficiency in large-scale, safety-critical environments. Experimental results demonstrate that the proposed RL-enhanced online learning unit significantly improves scheduling robustness and system efficiency. The adaptive capabilities of the online learning unit, driven by RL, ensure that the system can meet stringent timing constraints while dynamically adjusting to run-time variations, making it a highly effective solution for time-triggered architectures. This paper introduces a state of the art implementation that utilizes the use of machine learning with metascheduling pioneering the concept of AI to metascheduling in solving the scheduling problem.**

## I. Introduction

Meta-scheduling is frequently described in academic discussions as quasi-static scheduling [1], or as super-schedulers [2]. It is particularly crucial in safety-critical systems that need adaptive capabilities when a context event occurs (failure, slack or mode change) that renders the current operational schedule impractical or ineffective. Metascheduling in time-triggered systems plays a crucial role in adapting to dynamic and unpredictable conditions, ensuring both task execution reliability and efficiency. However, conventional methods face substantial challenges when attempting to train Artificial Intelligence (AI) -based scheduling models offline. This difficulty arises from the complexities of constructing a comprehensive Multi-Schedule Graph (MSG) that represents all possible scenarios. The process of developing an MSG that captures the extensive probability space—especially in the presence of context events such as hardware failures, slack variations, or mode transitions—requires significant resources and is often impractical. To address these issues, we propose the integration of an adaptive online learning module within the metascheduler, aimed at enhancing real-time performance. The driving motivation behind this approach lies in the limitations of offline training, which results in an MSG that is inherently a partial representation of the entire space, focusing primarily on the most likely and critical events. In contrast, the online mode utilizes Reinforcement Learning (RL) to continually explore and uncover novel scheduling solutions, thereby expanding the MSG and improving system performance over time. This dynamic adaptability enables the system to handle unforeseen events and intricate scheduling challenges more effectively. We implemented various RL models within the online learning module, each tailored to tackle specific scheduling issues. These models not only aid in discovering new solutions but also optimize existing schedules, particularly when stricter deadlines or new performance criteria emerge. By continually refining AI inferences through real-time training, the system remains flexible and capable of adapting to evolving demands, thereby ensuring robustness and efficiency in large-scale, safety-critical applications. Experimental results show that the proposed RL-enhanced metascheduler significantly enhances schedule resilience and system efficiency. The adaptive nature of the online learning module, driven by reinforcement learning, guarantees the system can meet stringent timing requirements while dynamically adjusting to runtime variations, positioning it as a highly effective solution for time-triggered systems.

This work expands on previous efforts to tackle the MSG state space explosion offline using machine-learning techniques by deeply exploring schedule variations through different levels of neighborhood aggregation and conducting a comprehensive comparison with previous methods cited in [3] [4], and [5]. While prior research has concentrated on addressing the MSG state-space explosion issue offline, this work builds upon that by tackling challenges that may arise in online mode. Specifically, it focuses on retraining machine learning inferences in order to prevent future missed deadlines. Such issues may stem from the inability to adequately explore the full range of failure probabilities and slack variations, as the high density and complexity of the MSG make it impractical to train for every possible scenario offline.

RL has been employed in this research to significantly enhance the scheduling framework used in distributed memory multi-core architectures. RL is adept at making sequential decisions, optimizing long-term rewards, and is particularly suitable for dynamic environments where scheduling decisions must adapt based on real-time system states and feedback. In the context of this research, RL has been utilized to optimize spatial allocation—the process of assigning tasks to specific processors in a way that balances load and minimizes communication overhead. Traditional spatial allocation methods often rely on heuristics that do not adapt to changes in workload or system configuration dynamically. In contrast, the RL-based model developed in this study learns from the environment through continuous interaction, adjusting its allocation strategies to maximize system efficiency and minimize computational delays. This integration of RL into the metascheduling process marks a substantial advancement in scheduling technology for distributed systems. It not only enhances the adaptability and efficiency of the scheduling framework but also reduces the energy and computational power required to maintain optimal system performance. The effectiveness of this model has been demonstrated through extensive simulations and real-world tests, which show significant improvements in task throughput and system responsiveness compared to traditional scheduling approaches.

The rest of the document is structured as follows: section II Mentions the related work for scheduling utilizing Deep learning, and re-enforcement learning in scheduling and Metascheduling emphasizing on the novelty of work in more details by explaining the additional inputs this research presents. Section III summarizes the theory behind the proposed workframe. Section IV explains the models and algorithmic techniques used to produce results, findings and methodologies are afterwards shown in section V, results are discussed in VI and the paper concludes with section VII.

## II. Literature review

Melnik et al. [6] proposed Neural Networks Scheduling (NNS), integrating Artifical Nueral Networks (ANNs) and RL for workflow scheduling. Their approach refines input states and optimizes makespan. Our research, in contrast, employs RL models to continuously enhance the performance of ANN schedulers and does not contribute directly to the scheduling process. Allowing faster execution time for the scheduling algorithm.

Murali et al. [7] propose a metascheduling algorithm to optimize job placement in high-performance computing (HPC) grids relying on day-ahead electricity markets, focusing on reducing electricity costs while ensuring efficient job scheduling. Their approach models the scheduling problem as a Minimum Cost Maximum Flow (MCMF) optimization and integrates queue waiting time and electricity price predictions to estimate execution costs. Through trace-based simulations using real-world workload traces and electricity price datasets, they demonstrate their method's ability to minimize electricity costs and optimize response times. However, their model assumes accurate price and queue predictions, which may not always be reliable, and does not account for unexpected events like hardware failures or network disruptions. Additionally, it overlooks makespan reduction, an important metric for some HPC applications, and may face computational challenges for large-scale real-time execution. In contrast, our work incorporates context events, safety checks, and makespan testing to ensure schedule quality.

Zhang et al. [8] introduced A Meta-Reinforcement Learning-Based Metaheuristic for Hybrid Flow-Shop Scheduling Problem with Learning and Forgetting Effects, a Q-learning-based metaheuristic for Hybrid Flow-Shop Scheduling (HFSP), optimizing makespan through dynamic operator selection. While both studies employ RL for scheduling, our work applies RL in a metascheduling context, prioritizing workload balancing, energy efficiency, and safety assurances against locked loops.

Liu et al. [9] proposed an actor-critic Deep Reinforcement Learning (DRL) model for Job Shop Scheduling, integrating convolutional networks to balance makespan and execution time. While both studies leverage RL, our approach targets real-time safety-critical systems, employing a metascheduler framework to dynamically adjust schedules based on contextual events while ensuring safety and precedence constraints.

Min et al. [10] introduced a metascheduling framework for 5G and beyond networks, leveraging a central unit for policy management and decentralized scheduling at base stations. Their RL-based cooperative learning approach enhances scalability and efficiency. Our work, while similarly using RL for dynamic scheduling, focuses on real-time safety-critical systems, incorporating workload balancing and energy efficiency.

Niu et al. [11] proposed a multiagent Meta Reinforcement Learning (Meta-RL) approach using Proximal Policy Optimi-

sation (PPO) for task scheduling in mobile-edge computing (MEC) environments. Their framework models scheduling as a non-cooperative stochastic game. While both studies optimize scheduling dynamically, our work integrates RL within a metascheduler to adapt to real-time context events, ensuring system reliability and adaptability.

Tang et al. [12] combined representation learning and DRL to optimize scheduling in Edge Computing, addressing scalability and dimensional challenges. Their approach maps tasks and nodes into vector spaces for efficient decision-making. Our work extends dynamic adaptation principles to metascheduling while considering workload balancing, energy consumption, and safety constraints.

Galstyan et al. [13] explored RL-based resource allocation in Grid computing using decentralized Q-learning agents. Their approach improves efficiency without direct communication among agents. Our research, while also leveraging RL for scheduling, focuses on real-time safety-critical systems, ensuring adaptability and reliability through Markovian properties and precedence constraints.

The paper by Liu et al. [14] addresses robustness and deadline guarantees in RL-based cloud task scheduling. It proposes a Meta-Learning-based RL framework to enhance adaptability in dynamic environments, reducing retraining time and improving deadline adherence. Our research similarly employs RL for scheduling but focuses on real-time safety-critical systems. While Liu et al. integrate metalearning for generalization, our approach incorporates a metascheduler for real-time context adjustments, prioritizing energy efficiency and task balancing alongside deadline adherence.

Suresh and Kumar [15] present an economic metascheduler for cloud ML services, leveraging decision trees to optimize algorithm selection and resource utilization. Both their work and ours employ intelligent scheduling to enhance resource efficiency. However, while their metascheduler selects optimal ML algorithms, our framework adapts schedules dynamically based on context events, addressing safety constraints such as precedence limitations and preventing locked loops.

Huang et al. [16] propose an RL-based scheduling approach for mobile edge computing systems with random task arrivals. Their model optimizes task offloading, transmission selection, and power allocation using RL in a Markov Decision Process (MDP). Our research similarly applies RL for scheduling but focuses on safety-critical systems, incorporating safety checks and ANN-based predictors to mitigate errors in decision-making. Unlike Huang et al., we address reconstruction and reliability constraints crucial for real-time applications.

Liu et al. [17] propose a meta-reinforcement learning framework to enhance RL-based scheduling robustness in cloud environments. Their approach integrates meta-gradient learning to improve adaptability, ensuring better deadline guarantees and reduced retraining time. While their focus is on time-critical task scheduling in the cloud, our research targets real-time safety-critical systems, incorporating precedence constraints and safety factors to prevent locked loops, enhancing system reliability under dynamic conditions.

Table I shows the summary of the comparative study between various works in the field of scheduling using reinforcement learning and other machine learning techniques. The final row represents work considered in this paper, which comprehensively addresses all the performance metrics summarized below:

- Scheduling: Indicates whether the work addresses scheduling of tasks.
- Metascheduling: Indicates whether the work includes metascheduling, which involves high-level scheduling strategies that adapt to different environments or contexts.
- Context Events: Indicates whether the work considers dynamic context events, such as hardware failures or schedule slacks.
- Safety check algorithms: Indicates whether the work includes algorithms to ensure safety, such as preventing locked loops or ensuring precedence constraints.
- Workload Balance: Indicates whether the work considers balancing workloads across resources.
- Makespan: Indicates whether the work aims to minimize the total completion time of all tasks.
- Energy Consumption: Indicates whether the work addresses optimizing energy consumption.
- Scalability: Indicates whether the work considers the ability of the algorithm to handle larger and more complex scheduling problems.
- Real-Time or Execution Time Consideration: Indicates whether the work accounts for real-time constraints or execution time considerations. This table provides a clear and concise comparison of the various works, highlighting the specific areas of focus for each research study and illustrating the comprehensive nature of our own work in addressing a wide range of performance metrics in scheduling.

This study presents an advanced implementation that integrates machine learning techniques within the domain of metascheduling, marking a significant advancement in the application of artificial intelligence to scheduling optimization. By pioneering the incorporation of AI-driven methodologies into metascheduling frameworks, this work seeks to address and enhance the efficiency of scheduling processes.

## III. Background

This section covers the required fundamental background theory to comprehend the proposed framework. Firstly, explaining scheduling and metascheduling, then explaining the theory behind RL models used in this research.

### A. Metascheduler and Multi-schedule Graphs

Fig. 1 shows the topological design of the metascheduler. It shows the Application Model (AM) which contains information regarding the precedence constrain, Worst Case Execution Time (WCET), and messages. Platform Model (PM) that contains information regarding the end systems, processing elements, links and routers. Context model (CM) which contains information regarding the possible context events such as failures, slacks or mode changes. The AM, PM and CM are fed into the metascheduler, which proceeds to compute the MSG. In our specific scenario, a Genetic Algorithm (GA) based scheduler is employed in offline mode to produce an MSG. The MSG is a Directed Acyclic Graph (DAG) that encompasses comprehensive information regarding schedules. In this graph, each node corresponds to a specific schedule, while the edges symbolise the transitions between schedules, which are influenced by context events as depicted.

TABLE I: Comparison of literature works

| **Work** | **Scheduling** | **Metascheduling** | **Context Events** | **Safety Checking Algorithms** | **Workload Balance** | **Makespan** | **Energy Consumption** | **Scalability** | **Execution Time Considerations** |
|---|---|---|---|---|---|---|---|---|---|
| [6] | X | | | | | X | | | |
| [7] | X | X | | | X | | X | X | X |
| [8] | X | | | | | X | | X | X |
| [9] | X | | | | | X | | X | X |
| [10] | X | X | | | | X | | X | X |
| [11] | X | X | | | X | X | | X | X |
| [12] | X | | | | | X | X | X | X |
| [13] | X | | | | | X | X | X | |
| [14] | X | | X | | | X | X | X | X |
| [15] | X | X | | | X | X | X | X | X |
| [16] | X | | | | X | X | X | X | X |
| [17] | X | X | | | X | X | X | X | X |
| **Ours** | X | X | X | X | X | X | X | X | X |

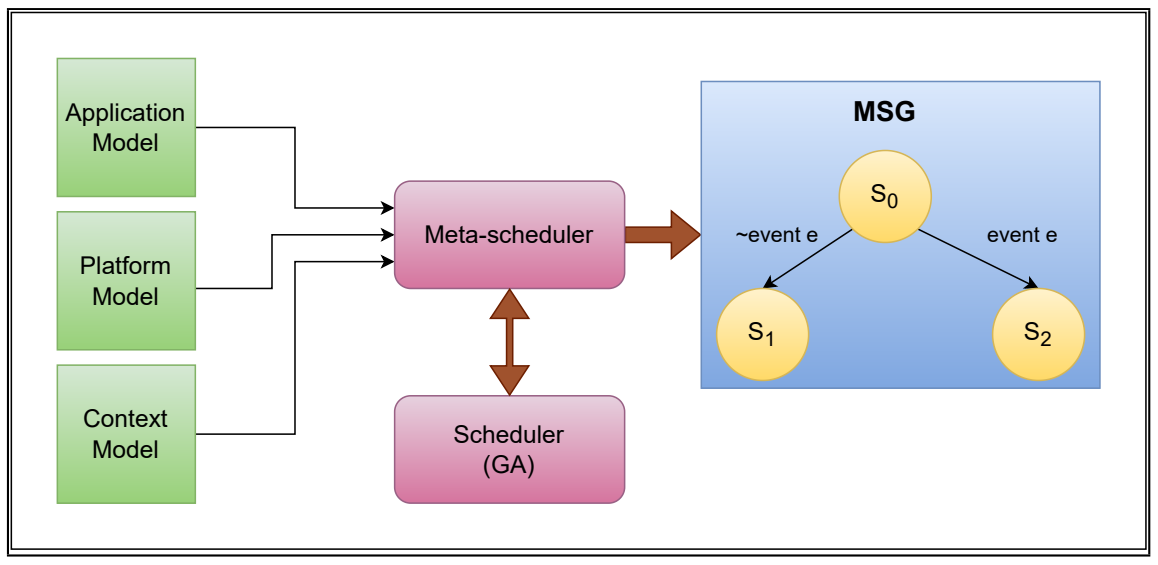

Figure 1: Metascheduler component [3]

The system employs an MSG, which is a DAG that represents precomputed schedules, designed to accommodate to different context events. Context events are events that causes the schedules to become ineffective or invalid such as hardware failures, schedule slacks and low power modes. in the MSG, schedules are denoted as nodes, and context event are denoted as edges. These schedules are dynamically switched during runtime in response to changes in context. AI inferences are trained offline by transforming the MSG into a dataset. The MSG contains information regarding schedules represented as nodes, and context events stored as edges in the DAG as shown in Fig. 2. $S_0$ represents the initial schedule, this schedule is employed in the system at the start and whenever a context event occures, whether its a failure in the hardware, a slack or a model change that causes the schedule to become ineffecient or invalid, another schedule is employed to take its place. $C_1$ for example is a slack event and $S_2$ is now employed. The solutions and new schedules are generated offline using the GA. The generated MSG covers a wide variety of schedules connected by edges. The MSG allows for the adaptation to context events. The aforementioned mechanism guarantees that the system complies with time-triggered constraints while adapting to new circumstances, since NN response rate is much faster than search algorithms classically used in literature, like GA [3] [4] [5]. Thereby preserving fault tolerance and system integrity. The metascheduling algorithm entails intricate decision-making processes that rely on the AM, PM, and CM. The approach mitigates the issue of potential state explosions by implementing reconvergence of paths and establishing reconvergence horizons within the MSG, effectively controlling the exponential increase in the size of the graph with increasing numbers of tasks and messages [18].

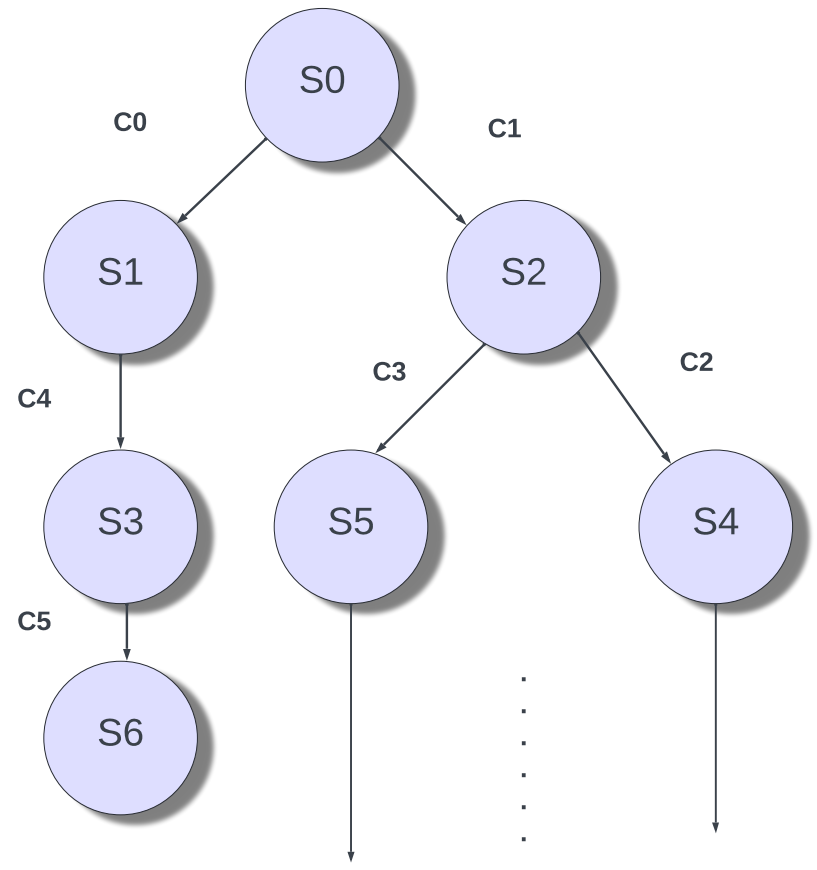

Figure 2: Fault handling in time-triggered systems

The primary challenge in expanding the domain of context events lies in the virtually infinite possibilities that arise when accounting for high-resolution slack variations, multiple failure scenarios, and mode transitions within the same CM. Consequently, only a subset of the MSG can be utilized for offline training of AI inferences to establish spatial and temporal priorities. As a result, the potential for errors during the system's online operation can never be entirely eliminated.

### *B. Reinforcement Learning and Online Learning Unit*

To mitigate the lack of MSG training data in offline mode, an RL-based model, termed the online learning unit, was developed to explore additional nodes in the MSG. RL was chosen due to its adaptability in interacting with the environment and it's unsupervised learning nature to further retrain AI inferences.

Multi-Armed Bandits (MAB) and Contextual Bandits (CB) illustrate core RL concepts, balancing exploration and exploitation. CB extends MAB by integrating contextual information to enhance decision-making. The update rule for CB follows Eq. 1:

$$Q_{t+1}(a) = Q_t(a) + \alpha_t(a)[r_t(a) - Q_t(a)], \tag{1}$$

where $Q_t(a)$ is the estimated value of action $a$ at time $t$, $\alpha_t(a)$ is the learning rate, and $r_t(a)$ is the observed reward [19]. This iterative process refines action selection over time.

Multi-Agent Reinforcement Learning (MARL) extends RL to scenarios involving multiple agents, where decisions impact both individual and collective rewards. MARL employs game-theoretic techniques to optimize policies. A fundamental approach is multi-agent Q-learning, which updates Q-values based on interactions among agents and environmental responses [20]. Challenges include credit assignment and dynamic agent behavior, requiring robust learning mechanisms to adapt effectively.

These RL approaches provide a scalable and adaptive framework for MSG exploration, leveraging iterative learning to enhance decision-making in uncertain environments.

It is crucial to acknowledge that our system has a Markovian property [21], indicating that the outcomes are independent of previous samples or time. This characteristic is beneficial when utilising the RL model equations previously described. RL algorithms provide substantial benefits in the domain of metascheduling, primarily because to their capacity for online learning and adaptability. Unlike traditional algorithms, RL has the capacity to continuously gain knowledge and adapt its tactics based on real-time input, removing the need for pre-labeled data. Online learning allows metaschedulers based on RL to dynamically adjust their scheduling decisions depending on changes in the environment or system state, hence enhancing their operational efficiency and effectiveness. Moreover, long-term optimisation of incentives is a feature of RL algorithms that makes them capable of managing intricate decision-making procedures frequently seen in assignment scheduling. Where workload and system resources may vary unexpectedly, flexibility is essential. RL can help adaptation and gradually enhance online AI inferences. As such, stronger and more resilient scheduling systems are produced.

## IV. System Overview and Implementation

This section presents a detailed implementation that underpins the proposed system architecture, explaining how different components interact within the framework. It details the structure of the system, emphasising the relationship between application, platform, and context models with a metascheduler that supports a variety of scheduling algorithms, including GNN, ANN, Encoder/ Decoder Neural Networks (E/D NN), and Random Forest Classifiers (RFC). The section also examines the function of online learning units, leveraging RL algorithms and neural network predictors to enhance adaptability in real-time scenarios. Additionally, it discusses reconstruction models that aid in the dynamic generation and assessment of schedules, ensuring the system's effectiveness in complex operational contexts.

### *A. System Overview*

Fig.3 presents a block diagram of the system architecture, highlighting the interplay between various components essential for its operation. The process begins with the input from the AM, PM, and CM, which are integrated within the information extraction block. This integration reshapes data to optimise its utility in the adaptation unit, ensuring that relevant information is readily accessible for decision-making processes within the Online Operation Manager. This decision-making hub evaluates context events to determine the appropriate response, such as activating spatial recovery reconstructors or invoking specialised schedulers focused on energy efficiency, minimal makespan, or workload distribution.

Outputs from the central unit include temporal and spatial priorities, alongside timing for context events, which are crucial for the subsequent scheduling process. These parameters are relayed to the reconstruction model, which undertakes the task of schedule formulation. This model incorporates mechanisms for handling message collisions, verifying precedence constraints, and ensuring safety, along with recordings of system components' execution times for robust recovery in response to context shifts.

Different reconstruction models are tailored to specific needs. Some reconstruct entire schedules, others focus on spatial priorities, and some are optimised for rapid response to contextual changes. The nuances of these models and their operational variances are explored further within this section.

The online learning unit leverages data from the information extraction block to initiate spatial allocation actions based on diverse reward structures. Feedback loops are established via observers that monitor system performance, adjusting rewards based on efficacy of actions taken based on profiles, whether its makespan, workload or energy consumption. As these actions inform the reconstruction blocks to generate updated schedules. The NN, initialized with weights from a pre-trained AI scheduling inference, is continuously retrained within the online learning framework. This framework employs reinforcement learning algorithms—such as Contextual Bandits, Multi-Armed Bandits, and Multi-Agent Bandits—to refine scheduling decisions. The NN dynamically adapts its parameters in response to real-time data or simulated context events (e.g., faults, slacks, or mode changes), ensuring continuous performance improvement even in the absence of actual events. through simulated fault injection scenarios as well as real senarios. The NN could be updated using enhancement learning to retrain scheduling AI inferences. The online learning unit is triggered in cases where the temporal or spatial priorities generated from the AI scheduling inference are not up to system standards (not reaching deadlines). Or in applications where there are no drawbacks by a continuously running online learning unit. In such cases the online learning unit would prevent the breach of deadline constrains in the future by retraining AI inferences after finding solutions that do not violate deadline constrains by deploying a RL based search algorithm.

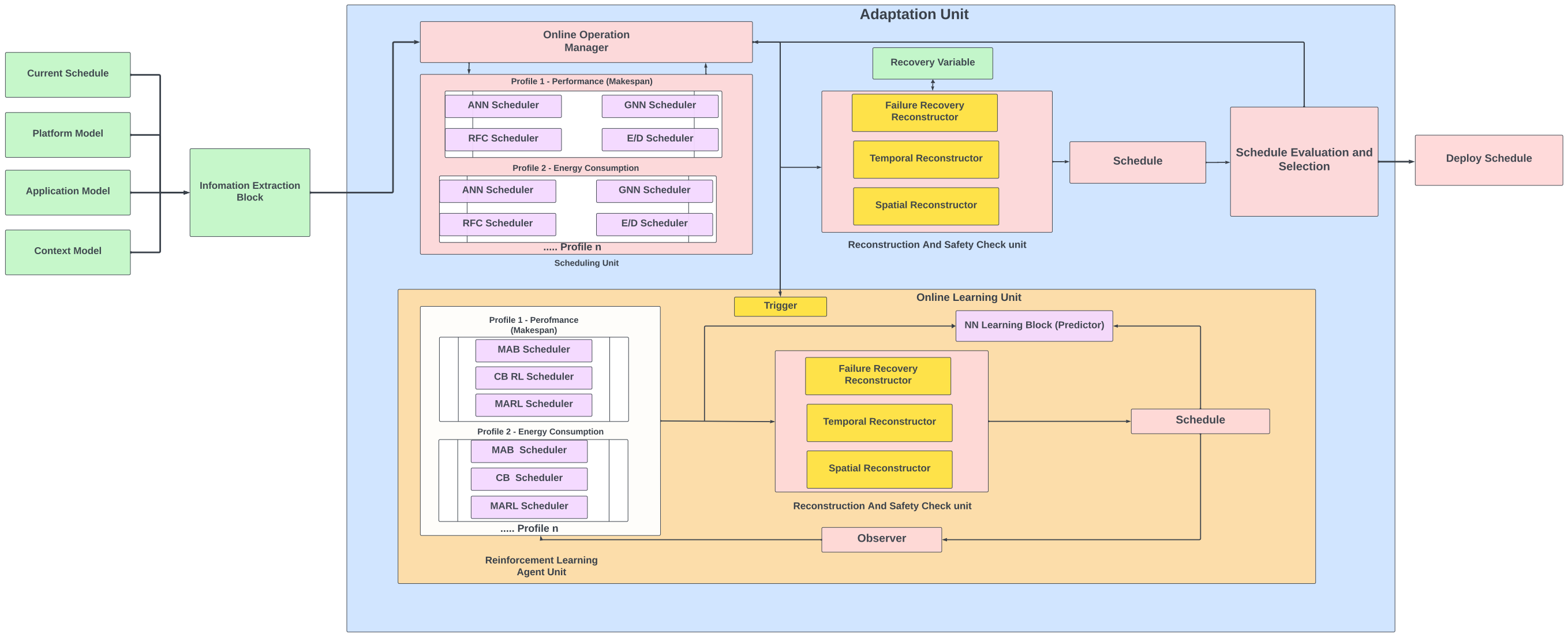

Figure 3: Complete system block diagram

### B. Context Handling

The fault management process within a Time Triggered System (TTS) is depicted in Fig.4, where the x-axis denotes the time steps ($t$) and the y-axis illustrates the end systems executing tasks (T1, T2, T3,... T8). The figure encapsulates two principal phases in context event management.

An intermediate scheduling phase is activated to mitigate the impact of the context event, illustrated by a system failure at End System 3 at $t_9$. This phase extends from $t_9$ to $t_14$, during which tasks continue to be executed in an adjusted schedule that compensates for the disrupted end system. Subsequent to this phase, at $t_14$ in Fig.4, a new schedule is implemented. This new scheduling framework accommodates the absence of End System 3 by reallocating tasks to alternative end systems, ensuring continuity. Ordinarily, this process is executed only when a failure occurs.

The system leverages distinct data packets extracted from the AM, PM, and CM, prepared within the information extraction block called consistency protocol, as shown in Fig.5. Bits 29 to 31 hold information regarding the context type, which could be a failure, slack or mode change, bits 26 to 28 hold the context value where the percentage of slack or power reduction is indicated. Bits 16 to 25 hold the location of a task affected by a context event. A timestamp for the context occurrence is stored in bits 6 to 15, and bits 0 to 5 store hardware IDs that are corrupted by failures. The consistency protocol is used to feed information to the online operation manager to aid in the decision making process.

This structured data provision supports the online operation manager in adapting to real-time requirements. For the generation of a revised schedule following a context event, the system utilises the CM to induce necessary modifications in the AM and PM. This approach is predicated on the nature of the context event: a slack affects the AM by altering task timings, a failure impacts the PM by adjusting resource availability, and a mode change may necessitate adjustments in both models. This dynamic integration ensures that the scheduling system remains responsive and adaptable to operational changes.

The mathematical representation of these interactions can be illustrated through Eq. 2, and Eq. 3, encapsulating the modifications induced by context events:

$$AM' = AM \oplus \Delta AM(CM_{\text{slack}}) \tag{2}$$

$AM'$ denotes the modified application model, $\oplus$ symbolises the update operation, $\Delta AM$ represents the changes dictated by the context model specific to slacks, and $CM_{\text{slack}}$ indicates the slack information within the CM.

$$PM' = PM \oplus \Delta PM(CM_{\text{failure}}) \tag{3}$$

Eq. 3 contains $PM'$, which is the updated platform model, $\oplus$ stands for the modification operation, $\Delta PM$ encapsulates the adjustments required by the platform model due to failures, and $CM_{\text{failure}}$ refers to the failure details provided by the CM.

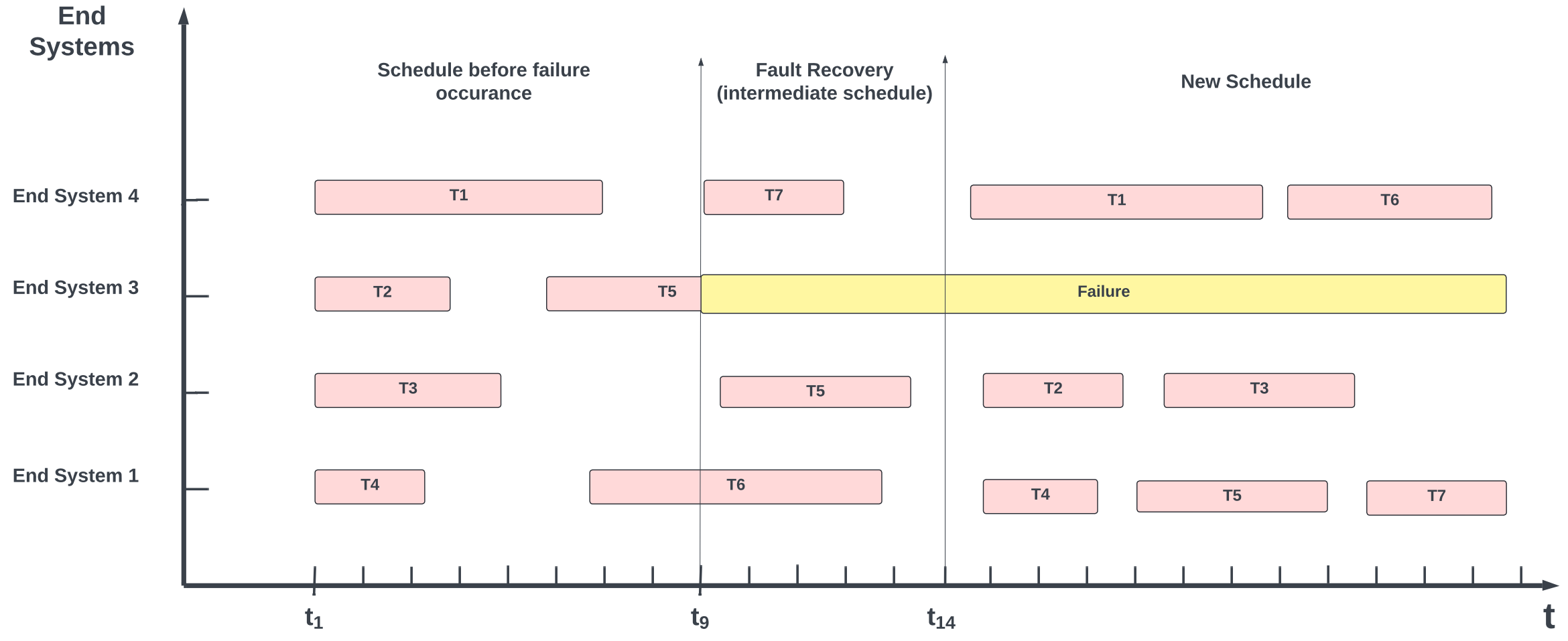

Figure 4: Fault handling in time-triggered systems

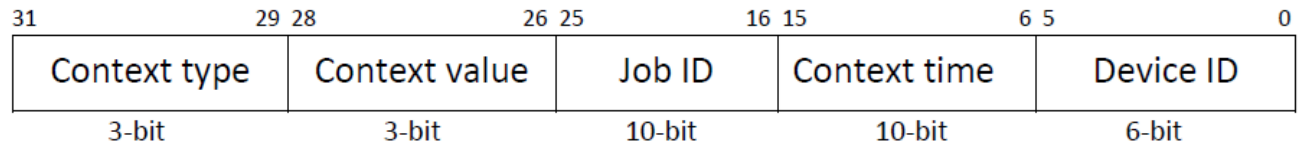

Figure 5: Consistency Protocol Design

These equations conceptualise how contextual changes are seamlessly integrated into the system's operational models.

### *C. Scheduling Unit*

The scheduling unit is tasked with processing inputs from the AM, PM, and CM. These inputs are then provided to the AI scheduling inferences, which use the extracted information to generate temporal and spatial priorities. The AI inferences play a crucial role in determining key parameters to generate schedules by analysing the input data and producing priorities that guide the execution order and resource allocation for tasks within the system.

As depicted in Fig.6, the operation of the scheduling unit is initiated by the online operation manager, which is responsible for aggregating crucial data from the AM, PM, and CM through the system information gathering block. This block serves as the interface between the different models and the AI scheduling inferences, ensuring that the most relevant information is extracted for processing.

The collected data from the AM, PM, and CM encompass various parameters that reflect the current state of applications, platform resources, and contextual conditions. This information forms the input to either the temporal AI inferences or the spatial AI inferences, both of which are trained offline using the MSG. The training process equips these inferences with the capability to analyse input data and generate temporal or spatial priorities, tailored to optimise the scheduling process.

Once the AI inferences have processed the input data, they generate temporal and spatial priorities, which are essential for determining the order of tasks execution and the allocation of resources. These priorities are then passed on to the reconstruction model, which synthesises them into schedules. The reconstruction model integrates these priorities to form a holistic schedule that ensures effective task execution, taking into account the dynamic nature of the system's operational environment.

### *D. Reconstruction Models*

The reconstruction model plays a pivotal role within the overarching system architecture, serving as a critical component that translates the decisions made by the scheduling inferences into actionable schedules. The reconstructor takes the outputs from the online operation manager, which include temporal priorities, spatial priorities, and context event timings, to assemble a coherent and executable schedule. It ensures that all operational constraints, such as message collisions and precedence requirements, are met while also integrating safety checks.

The design and functionality of different reconstruction models vary; some are tailored for comprehensive schedule rebuilding, while others focus on rapid response to immediate contextual changes. This section delves into the various aspects of reconstruction models, detailing their integration with the system's dynamic environment and their critical role in maintaining system resilience and adaptability.

The primary components of the reconstruction model are depicted in Fig.7. The generation of scheduling information, which includes temporal and spatial priorities, necessitates the integration of the AM, PM, and CM. Priorities may be derived from basic built-in algorithms or more sophisticated sources, such as ML models or GA. The built-in algorithm uses bottom level to determine the temporal priorities of a task, taking into consideration WCET and their precedence constraints, as shown in Eq. 4:

$$b\text{-level}(v) = w(v) + \max_{(v \to u) \in E} (b\text{-level}(u)) \quad (4)$$

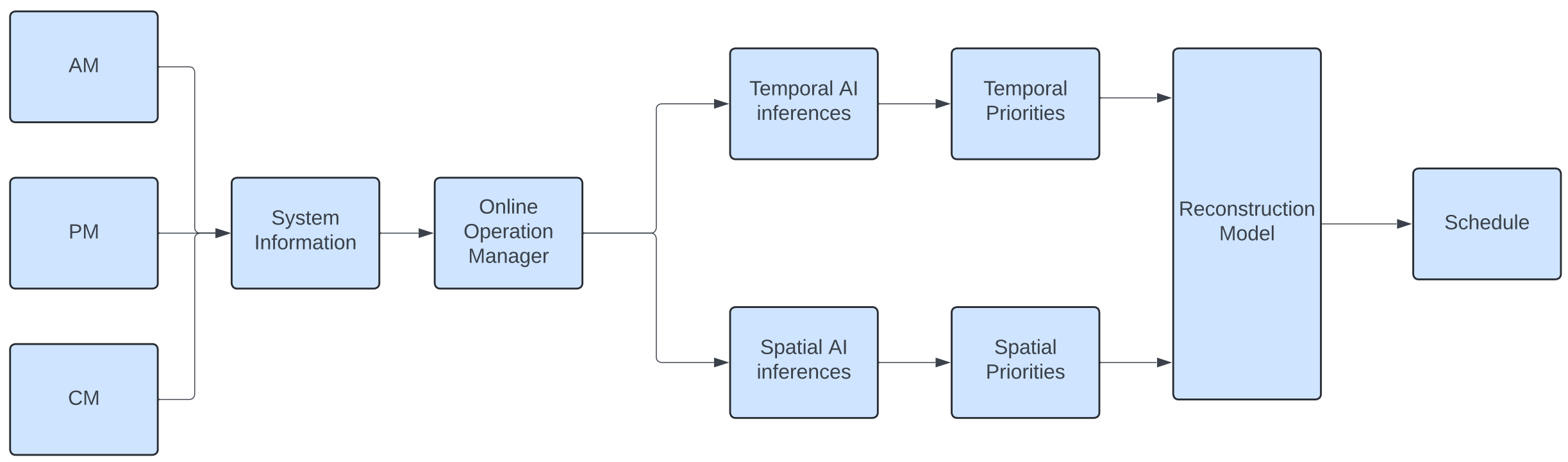

Figure 6: Component of the scheduling unit

Where:

- $b$-level$(v)$ represents the bottom-level value of task $v$, which is an estimate of the longest path from task $v$ to any exit task in the graph.
- $w(v)$ denotes the execution time or the weight of task $v$.
- $\max_{(v \to u) \in E}(b\text{-level}(u))$ calculates the maximum b-level among all tasks $u$ that are direct successors of $v$ in the task graph, where $E$ represents the set of edges indicating dependencies from task $v$ to task $u$.
- Task Weight $(w(v))$: This is the computation time required to complete the task $v$.
- Precedence Constrains: The maximum b-level among all successors ensures that the longest path from the task $v$ through any of its successor paths is considered [22].

As for spatial priorities, when the built-in algorithm is employed, it uses a search algorithm for the least loaded end system to deploy tasks.

In the event of a context that necessitates fault recovery, relevant details such as the event timing, affected end systems, affected tasks, and modifications to the WCET are extracted from the CM.

Recovery variables are stored from the reconstruction model, which logs the internal variables of the reconstructor over time. These variables are loaded at the moment hardware failure occurs, fixing tasks executed prior to the context event and adjusting information for tasks occurring after the context event, saving significant time caused by the re-computation of new variables in the case hardware failure occurs.

For the generation of new schedules $AM'$ and $PM'$ are extracted, which are derived by integrating updates into the original AM, PM, and CM. These modifications reflect changes induced by context events, transforming the original AM and PM accordingly. These transformations are detailed in Eqs. 2, and 3 demonstrating how dynamic context elements systematically influence the scheduling framework.

The schedule reconstructor performs several critical functions:

- **Tasks Allocation**: assigns tasks to appropriate end systems based on priority information and system capacities.
- **Message Allocation**: manages data transfers between tasks while preventing message collisions.
- **Fixing Past**: in fault recovery scenarios, fixes the parameters of tasks executed prior to the context event and reallocate only those tasks that are pending, as detailed in previous sections.
- **Safety check**: the reconstructor ensures precedence constrains and handles message collisions.
- **Schedule Generation**: ultimately, it constructs an update schedule.

Finally, an evaluation block is employed to generate performance metrics tailored to assess the schedules based on various profiles and evaluation criteria. When changing profiles, essentially, this is the only block that changes in the reconstructor. The evaluation block would calculate either makespan, energy efficiency or workload of the produced schedule.

### *E. Online Learning Unit*

This section delves into the online learning components, focusing on the deployment of RL algorithms such as CB, MAB, and MARL. These algorithms enhance the adaptability and performance metrics of the system by dynamically optimising decision-making processes in response to real-time data. The online learning unit, as outlined in the system overview, employs these algorithms to continuously refine and adjust the system's operations based on incoming feedback and context changes. This provides continuous learning for the AI inferences in the face of variable conditions and leverages the unsupervised learning power of RL to anticipate and mitigate potential disruptions. To fully comprehend the functionality of an online learning unit, the following aspects are discussed separately:

*1) Online Learning Unit Motivation:* The primary motivation for developing the online learning unit arises from the significant challenges associated with training AI scheduling inferences offline using the MSG generated from the GA. The process of creating an MSG that comprehensively accounts for all possible scenarios is exceedingly resource-intensive, given the vast probability space involved. For instance, when considering hardware failures as context events, the system must account for a wide range of potential failures—such as the failure of a link, router, end system, or various combinations of these elements. Furthermore, when slack occurs in the scheduling process, the degree of slack can vary anywhere from 0% to 99% for each scheduling task. If the context event involves a mode change, such as dynamically adjusting the frequency to reduce energy consumption, the range of possible adjustments is virtually limitless. These complexities are further exacerbated in large-scale scheduling scenarios where the number of tasks is substantial, making the construction of a complete MSG extraordinarily difficult, if not impossible.

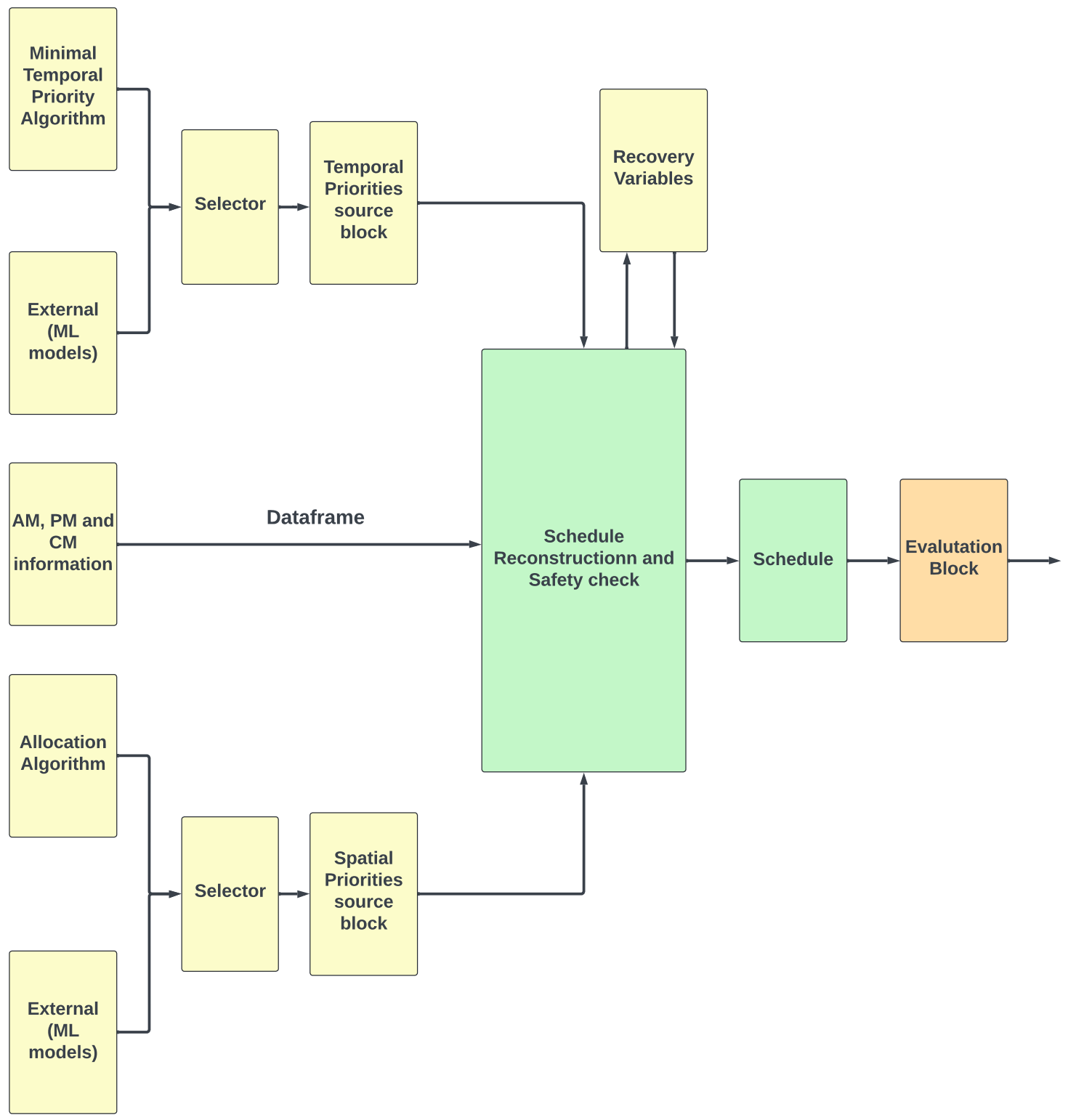

Figure 7: Building blocks of reconstructors

To address this challenge, the MSG created in offline mode is designed as a subset of the complete MSG, focusing on the most probable and critical context events. This approach allows for efficient use of resources while ensuring that the system is prepared for the most likely scenarios. During the online mode, the RL component continues to explore and discover new solutions, potentially expanding the coverage of the MSG. As the RL algorithm operates, it trains the AI inferences in real-time, progressively enhancing the system's performance over time, as illustrated in Fig.8. Which provides a visual representation of this concept, where nodes represent schedules and edges denote context events. The blue nodes indicate the schedules that were included in the offline training dataset, having been trained and prepared for likely scenarios. In the online mode, the online learning unit actively induces context events, which leads to the discovery of new nodes within the MSG—nodes that were not part of the initial offline training dataset. This continuous process of exploration and learning allows the system to adapt dynamically, improving its ability to handle unexpected events and complex scheduling scenarios as they arise.

Additionally, the online learning unit can be leveraged to explore improved solutions for existing schedules, particularly in scenarios where the current solutions fail to meet specific deadlines. For instance, if a stricter deadline is introduced, the online learning unit can actively search for alternative scheduling strategies that better align with the new requirements. Moreover, it can enhance the performance of the solutions generated by the AI inference through continued training and refinement. This process, known as enhancement training, allows the AI inference to adapt to evolving demands and constraints, thereby optimising the scheduling outcomes even further. By continuously refining its approach, the online learning unit ensures that the system remains flexible and capable of meeting more stringent performance criteria as they arise.

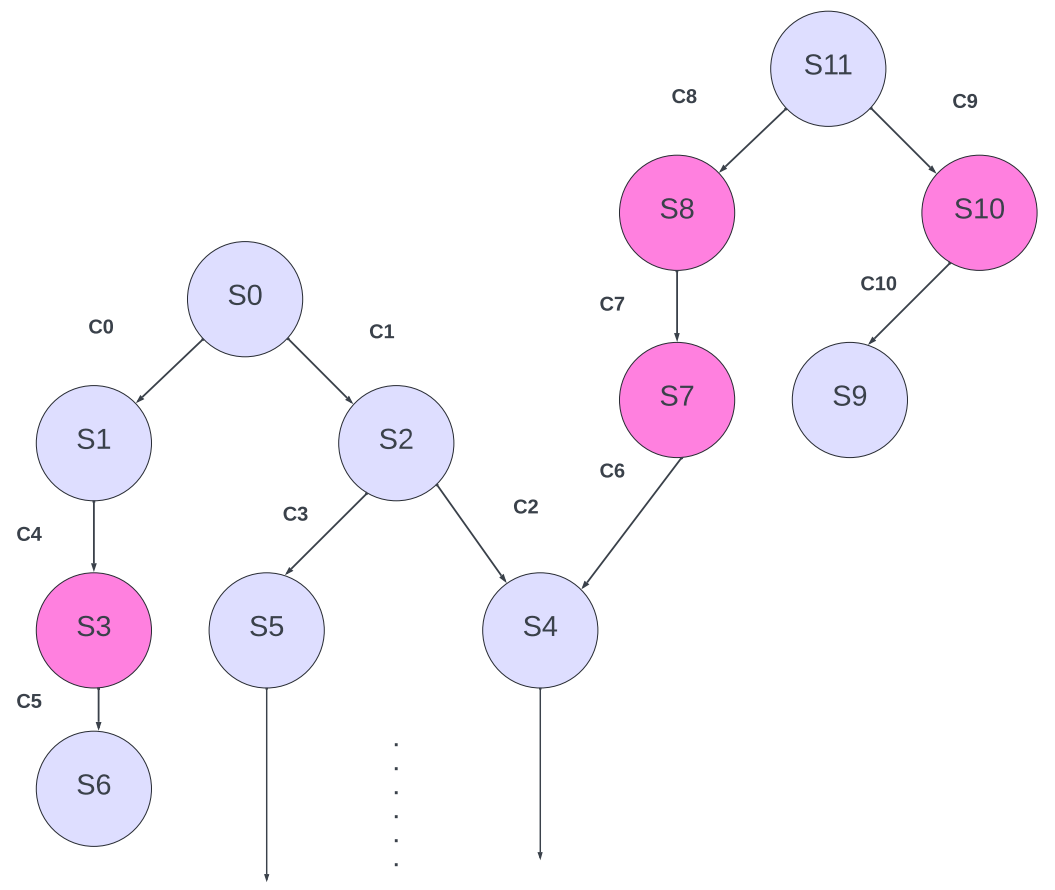

Figure 8: Multi-schedule graph training nodes

*2) Triggering Online Learning Unit:* The online learning unit requires high computation power leading to high power consumption that potentially affects the system performance. The activation of the online learning unit must meet specific requirements and must be activated under one of two conditions:

(a) If the priorities generated by the AI inferences lead to schedules that fail to meet the system's deadline requirements, as determined by the schedule evaluation unit, the system will proceed to further train the AI inferences. Once the target deadline is achieved, it will update the weights.

(b) If the application permits continuous learning, when continuous processing is not a primary drawback to the system. In such cases, the trigger can be permanently "on" to allow the system to continuously enhance its performance through enhancement training.

*3) Enhancement Training:* Online learning unit is considered beneficial to the adaptation due to the unsupervised learning capabilities that are utilised to train NN. This feature can be used to enhance the performance of AI scheduling inferences in generating temporal and spatial priorities that leads to better schedules in terms of makespan, workload distribution or energy efficiency. The AI inferences utilised in the scheduling unit can also be leveraged within the online learning unit. When the online learning unit is activated, the parameters of the AI scheduling inferences can be transferred to a NN predictor. Further training on these parameters allows for refinement and improvement. Once the parameters are updated, they can be sent back to the AI scheduling inference by copying the newly modified parameters into the AI scheduling model.

In this setup, the NN predictor serves as a proof of concept, demonstrating that NNs can effectively learn from the unsupervised decision-making processes within the online learning unit. However, the process of retraining and enhancing the AI inferences is inherently more complex. Given hardware limitations, this may necessitate conducting the enhancement training process externally, such as by transferring the parameters to a edge-based system and then integrating the modified parameters back into the local system or using a dedicated hardware with an external clock that is much faster than the regular system. However, it should be noted that the search algorithm chosen for the online learning enhancement unit is completely parallelable, meaning if hardware resources are available it would be possible to execute the algorithm in one time cycle. Moreover, while enhancement training can improve inference accuracy, it also introduces the risk that modified parameters could lead to worse decisions when applied to scenarios similar to those previously encountered by the AI inference during offline training. Which is why it is necessary to carefully select the accuracy testing samples and only commit the model once the testing accuracy improves. This potential for generating less efficient schedules underscores the importance of carefully managing the retraining process. Hence, the need for advanced training techniques must be carefully studied and thoroughly researched to ensure inference effectiveness and reliability.

*4) Multi-Arm Bandit Design:* Fig.9 illustrates the architecture of the MAB. Within this framework, MAB utilises a set of options, referred to as arms, to propose solutions or actions. These are governed by a policy—specifically, a refined version of the epsilon-greedy algorithm incorporating epsilon decay. This modification enhances the strategic balance between exploration, the pursuit of new possibilities, and exploitation, the utilisation of the best-known actions derived from accumulated rewards. As operational time extends and solutions approach optimality, epsilon decay progressively reduces the epsilon value, thus shifting the focus from exploration to increased exploitation. Actions determined by this policy are processed by the reconstruction model to create schedules, which are then evaluated for performance metrics (rewards) by the observer. The best schedules are catalogued for future reference. The reward structure varies according to the operational profile, which may prioritise objectives such as energy efficiency, performance optimisation, or workload distribution.

*5) Contextual Bandit Design:* Fig.10 delineates the design of the CB, which integrates input features into its decision-making process, in contrast to the MAB. These input features are derived from the PM, AM, and CM. The incorporation of the CM modifies the role of the observer; it no longer solely associates actions with rewards but also aligns actions with specific contexts to accommodate varying outcomes across different situations. Unlike the MAB, which utilises multiple arms, the CB design simplifies to a single decision-making process informed by context. In addition to the basic CB model, a NN predictor is incorporated within the online learning framework to simulate and study the system's dynamic behaviour. This NN predicts rewards, essentially mirroring system operations. During the RL phase, the NN training component may interface with various data streams. This setup facilitates transfer learning, allowing the utilisation of weights from AI-driven scheduling inferences to augment the metascheduler's decision-making capabilities. Periodically, these weights are reciprocated to refine their precision, thus enhancing overall schedule generation process.

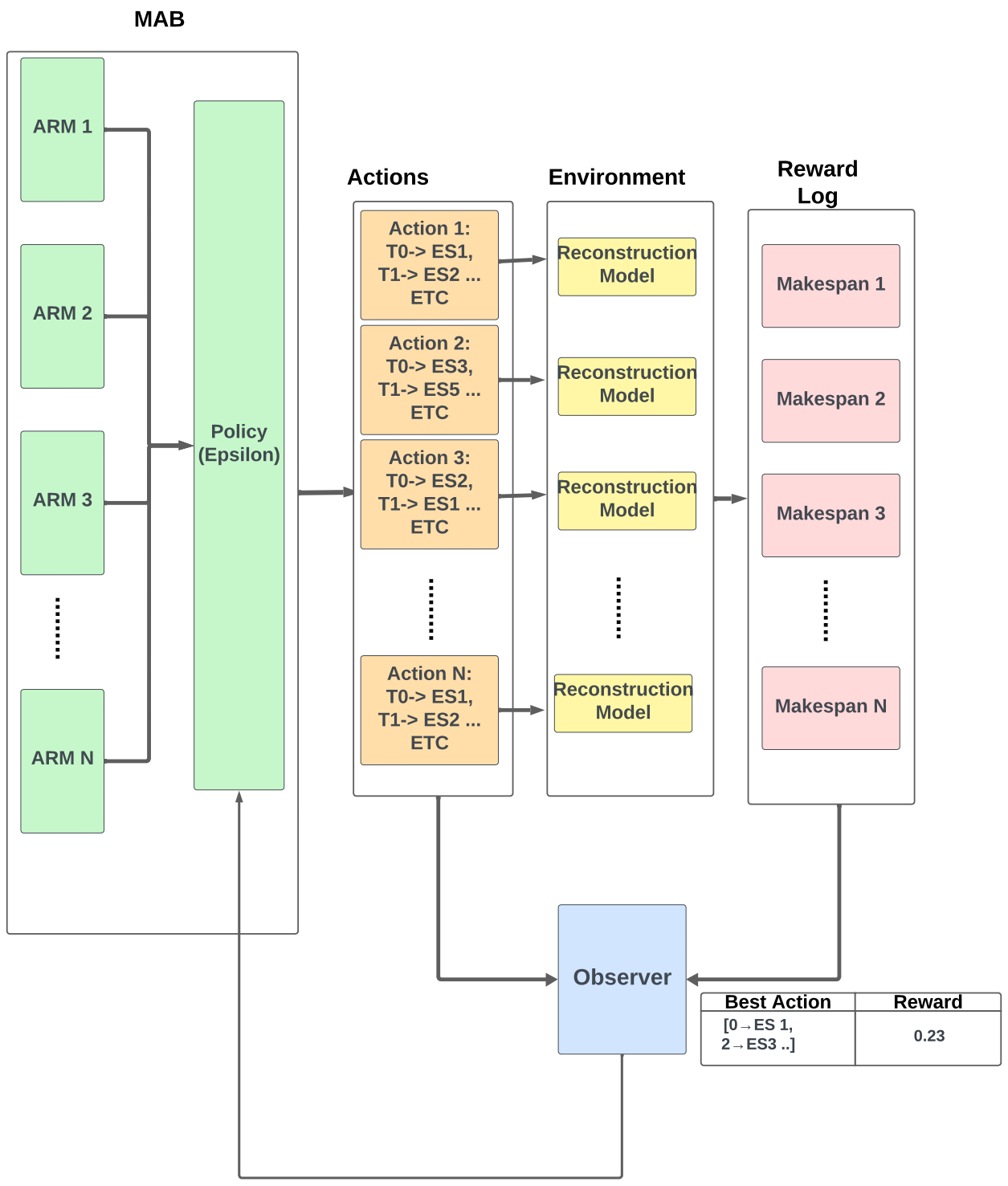

Figure 9: The Multi Arm Bandit Design

*6) Multi Agent Reinforcement Learning Model Design:* Fig.11 illustrates the design of the MARL system. In this architecture, each task is managed by a dedicated agent responsible for determining its respective temporal or spatial priority. A distinctive feature of this setup is the inclusion of a coordinator agent, which aggregates the decisions from individual agents into a unified strategy. This approach enhances decision-making efficiency as each agent autonomously evaluates the best course of action based on specific attributes, thus optimising the overall system response. Beyond these enhancements, the remainder of the system's components continue to function as previously established. Overall, the environment referred to in RL maps to the reconstruction model and the generation of schedules.

### *F. Time-Triggered System Overview*

In TT architectures, as shown in Fig.12. The system timeline illustrates the sequence of operations that occur when the various components of the system function together in response to context events. The process begins when a context event, such as "Event X," triggers the system. The first step involves invoking the AI inference, which is responsible for generating temporal and spatial priorities based on the current system state.

Once the priorities are generated, the system proceeds to reconstruct the schedule and perform a safety check to ensure that the new schedule meets all required constraints. If the end of the schedule cycle is reached and a missed deadline is detected, the system triggers the Online Learning Unit. This unit is tasked with exploring new scheduling solutions that may better meet the stringent deadlines or other performance criteria.

The timeline also indicates how the system responds to subsequent context events (e.g., "Event Y"). After the AI inference generates priorities and reconstructs the schedule, the system checks if the new schedule meets the deadlines. If a context event like "Event X" occurs again, the system modifies the previously generated schedule (S1), ensuring it is updated to account for the new conditions.

The exploration stage follows, where the RL model takes initial actions to explore potential solutions. The system continuously balances exploration and exploitation to detect better solutions. If a superior solution is found, the system commits to this solution and transfers the modified AI inference parameters back into the system, ensuring that the enhanced schedule is now part of the operating model.

## V. Methodology and Experimental Setup

This section outlines the tools used to design and test the system, detailing techniques for data collection, analysis, and validation. Additionally, it describes how data changes throughout the system and It describes parameter tuning and evaluation to ensure replicability and reliability.

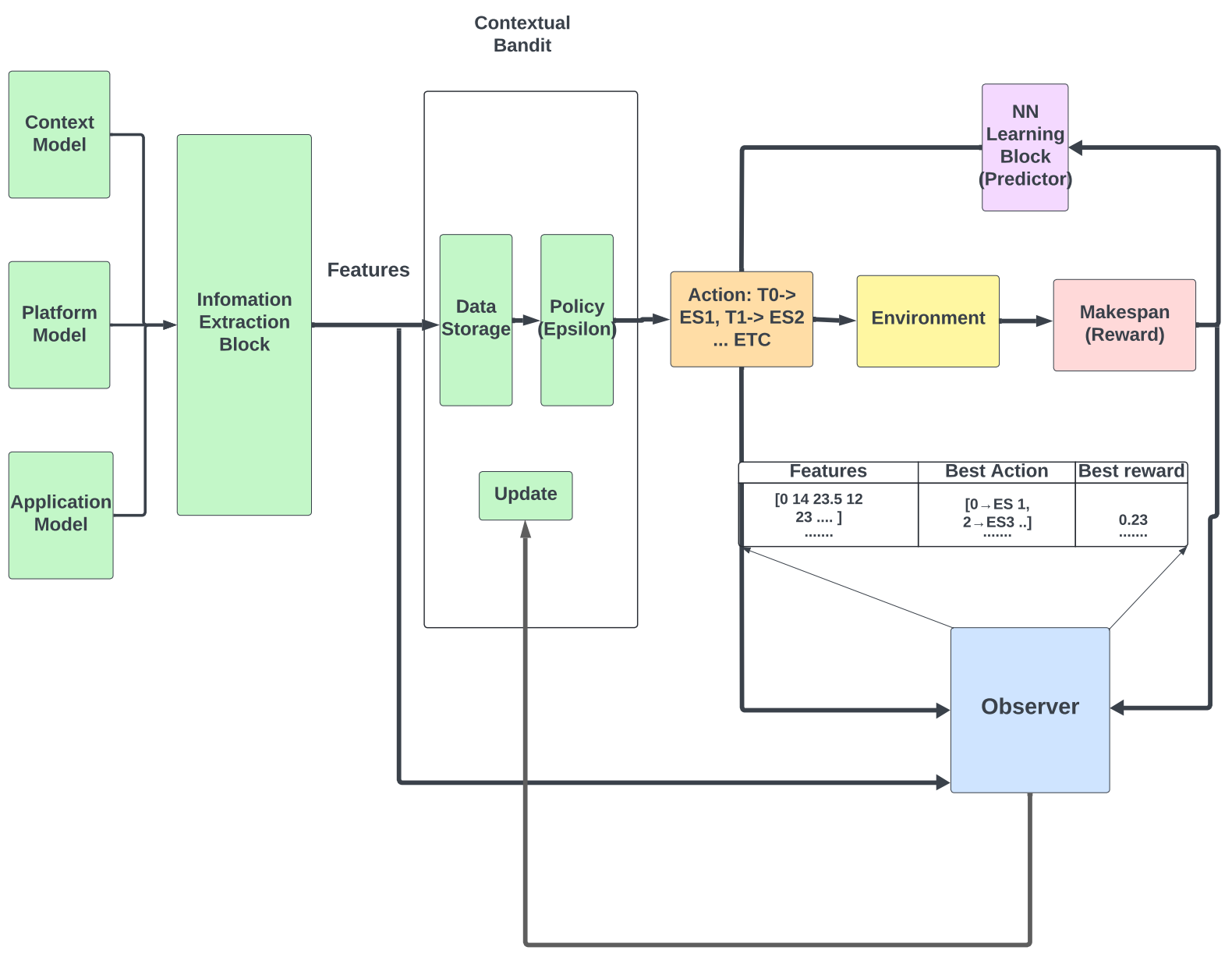

Figure 10: The Contextual Arm Bandit Design

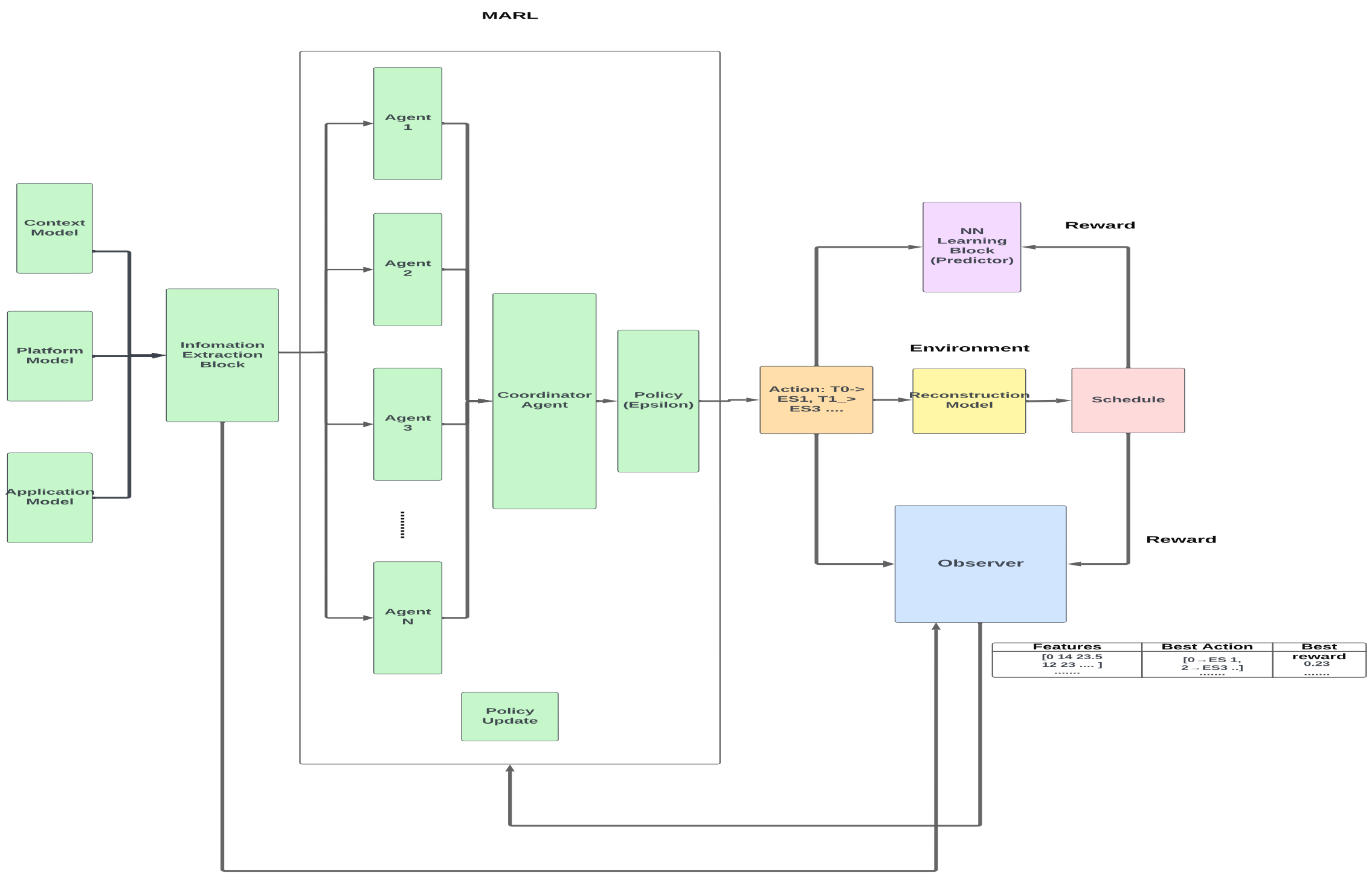

Figure 11: The Multi Agent RL Model Design

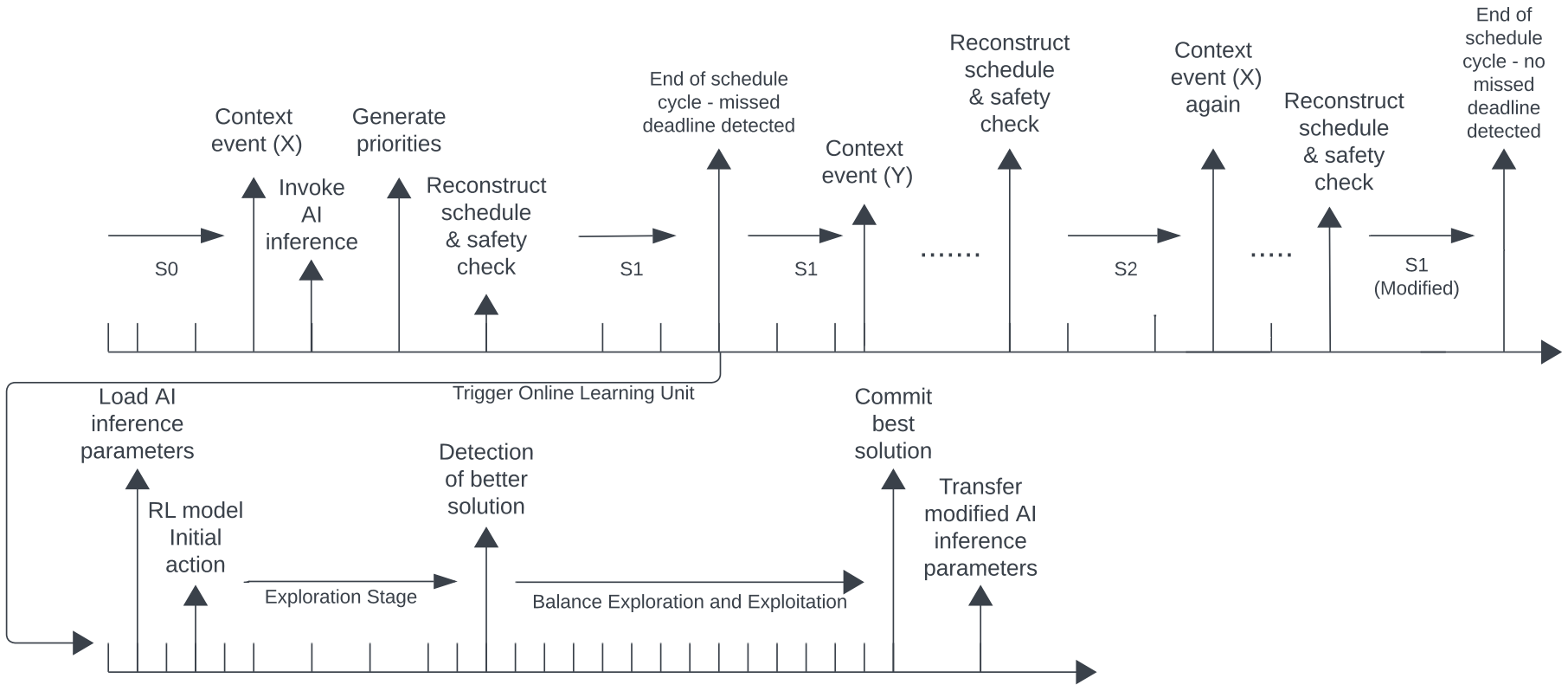

Figure 12: Timeline showing how system component work

## A. Libraries and Tools

Several libraries and tools were used for simulations, experiments, and data analysis, primarily in Python. Key libraries include:

- **PyTorch** [23]: ML library for neural networks.
- **PyTorch Geometric** [24]: Deep learning on graphs.
- **NumPy** [25]: Multi-dimensional arrays and mathematical functions.
- **Scikit-learn** [26]: ML algorithms for classification, regression, and clustering.
- **Snap** [27]: High-performance network analysis.
- **NetworkX** [28]: Graph creation and analysis.
- **Google Colab** [29]: Cloud-based Jupyter environment.
- **Matplotlib** [30]: Data visualization.

These tools facilitated efficient data processing, model training, and evaluation, supporting the research objectives.

## B. Dataframe Evolution in the System

This section outlines the transformation of data within the system, focusing on the Task Dataframe and Message Dataframe, both structured as Python dictionaries. These dataframes progress from an initial empty state to a fully populated state after undergoing Application and Platform Model processing, Scheduling Inferences, and Reconstruction.

Initially, the Task Dataframe stores essential task-related attributes such as Task ID, WCET, dependencies, temporal priority, execution system, and scheduling times as shown in Table III, while the Message Dataframe captures inter-task communication details, including message identifiers, transmitting and receiving tasks, and timing information. As the system processes the data, these dataframes are incrementally populated, integrating scheduling inferences and reconstruction mechanisms to achieve a fully scheduled system.

Upon completion, the dataframes provide a comprehensive schedule outlining precise task execution and communication timing, ensuring predictability and reliability in real-time systems as The final state of the system is represented by Table IV. :

TABLE III: Input Features for Datasets

| Name | Description |
|---|---|
| *task_id* | Unique identifier for each task |
| *runs_on* | Identifier for the machine executing each task |
| *start_time* | Commencement time of each task |
| *wcet* | Worst Case Execution Time for each task |
| *msg_id* | Identifier for messages between tasks |
| *inj_time* | Time at which messages are injected |
| *msg_size* | The size of the message |
| *ridx* | Router index indicating the start and subsequent hops |
| *context* | 32-bit information related to context events |

TABLE IV: Final State of Message Dataframe

| Message ID | Tx Task | Rx Task | Tx End System | Rx End System | Start Time | End Time |
|---|---|---|---|---|---|---|
| 1 | 1 | 2 | ES1 | ES2 | 10 | 15 |
| 2 | 1 | 3 | ES1 | ES3 | 10 | 20 |
| 3 | 2 | 4 | ES2 | ES4 | 25 | 30 |
| 4 | 3 | 5 | ES3 | ES5 | 30 | 35 |

## C. Hyperparameter Tuning

Hyperparameter tuning is essential for optimizing ML models, balancing performance and generalisation. Common techniques include grid search, random search, and Bayesian optimisation. Grid search exhaustively evaluates all hyperparameter combinations, ensuring optimal selection but at high computational cost [31]. Random search offers a more efficient alternative by randomly sampling configurations, often yielding good results faster [32]. Bayesian optimisation models performance probabilistically, guiding exploration towards promising configurations [33].

For this research, various RL models were fine-tuned:

- **Multi-Armed Bandit:** Epsilon starts at 1 and decays to 0 with a discount factor of 0.963.
- **Contextual Bandit:** Epsilon decays with a 0.96 discount factor; ANN with two hidden layers (10 neurons each); 100 iterations and a 0.001 learning rate.
- **Multi-Agent RL:** Epsilon decays with a 0.99 discount factor; 16 neurons per layer; 0.1 learning rate.

These configurations ensure efficient learning and adaptability across distributed tasks and environments. The parameters were tested and selected using experiments detailed in the next sections.

### *D. Experimental Setup*

The online learning unit, integral for adaptive scheduling, is scrutinised to evaluate its performance over time and across different operational scenarios. This includes tuning of RL parameters and analysing the learning convergence to optimise system responsiveness and efficiency. The tests were conducted on 3 RL algorithm models: the MAB, CB, and MARL.

Detailed experiments designed to optimize RL algorithm parameters, such as epsilon decay, are conducted. The focus is on determining the optimal settings that balance exploration and exploitation, thereby enhancing learning efficiency and system performance, as well as experiments to study the complexity online learning components. This experiment assesses the computational demands of the online learning components, analysing the time efficiency and resource utilisation across different RL models and task configurations.

Fig.13 shows the monitoring component in the online learning unit. As the NN predictor was monitored for prediction error, while a reward monitor was installed to determine the reward generated from the spatial priorities generated from the RL inferences after reconstruction. A time monitor was also installed to determine the time taken for the inference to make a decision.

Actions created in the online unit are spatial allocation decisions. The experiment also considers allocations based on profiles, where makespan, workload and energy efficiency are taken into consideration as reward functions.

Additionally, the rewards of the system are monitored with every time step (epoch) to determine the learning capabilities of each algorithm. Essentially, the goal of the RL algorithm is to continuously create decisions that improve schedule performance (reduce makespan, distribute workload or increase energy efficiency) and to insure that the NN predictor is learning while the RL is operating. Thus, increase in rewards is monitored with epochs and NN prediction error is also monitored with time.

## VI. Results and Discussions

This section shows the results obtained from the previously explained experiments. The results are analysed and discussed in this section, afterwards, they are organised to evaluate the performance of scheduling inferences, online learning unit, and adaptation component respectively.

The experiment required for the MAB model to create spatial allocation actions for different number of tasks. The experiment held 100 different scenarios with 10 different context events for each example. The purpose of the experiment was to determine the proper epsilon decay value to balance exploration and exploitation as shown in Fig.14. Where multiple trials were conducted to determine the most appropriate epsilon decay values.

After selecting the proper epsilon decay value, the reward are obtained are observed with each episode to generate the exploration and exploitation graph as seen in Fig.15. it is noticed that the graph continues exploring and slowly settles in to the maximum value at episode 500 where epsilon values goes into full exploitation mode at episode 700.

Fig.16 shows the difference in performance between each RL model with increasing number of tasks. It is noticed that in applications that require low number of tasks, all models have similar performance which is understandable due to the fact that with low complex scheduling tasks it would more likely to converge to the best action. However as the scheduling problem becomes more complex the difference in performance becomes more clear, especially with the MARL model as it produces solutions that have much higher reward. Fig.17 shows a summary of max obtained rewards vs to the number of tasks for each RL model.

However, the enhanced rewards obtained from each model come with the trade-off of increased time consumption and higher processing demands, as demonstrated in Fig.18. Based on the analysis of the runtime data, the computational complexity of the three decision-making algorithms is derived. The MAB algorithm exhibits a linear complexity of $O(n)$, indicating that its runtime increases proportionally with the number of jobs. In contrast, the CB algorithm follows a quadratic complexity of $O(n^2)$, suggesting a significantly higher computational cost as the problem size grows. Meanwhile, the MARL algorithm demonstrates a logarithmic complexity of $O(\log n)$, making it the most scalable among the three. These results highlight the trade-offs between computational efficiency and decision-making strategy across different algorithms. It is evident that the MARL model incurs significantly longer execution times compared to the CB and MAB models. While the CB model does have a slightly higher execution time than the MAB model, the difference in their performance is not substantial. This marginal difference suggests that the CB model's increased execution time does not justify its use over the MAB model, making the CB model a less practical and feasible solution despite its marginally better performance. The considerable processing time required by the MARL model, although leading to higher rewards, presents a significant drawback in scenarios where computational efficiency and time constraints are critical. Therefore, while each model offers distinct advantages in terms of rewards, their respective time and processing requirements must be carefully considered in determining their overall feasibility for practical applications.

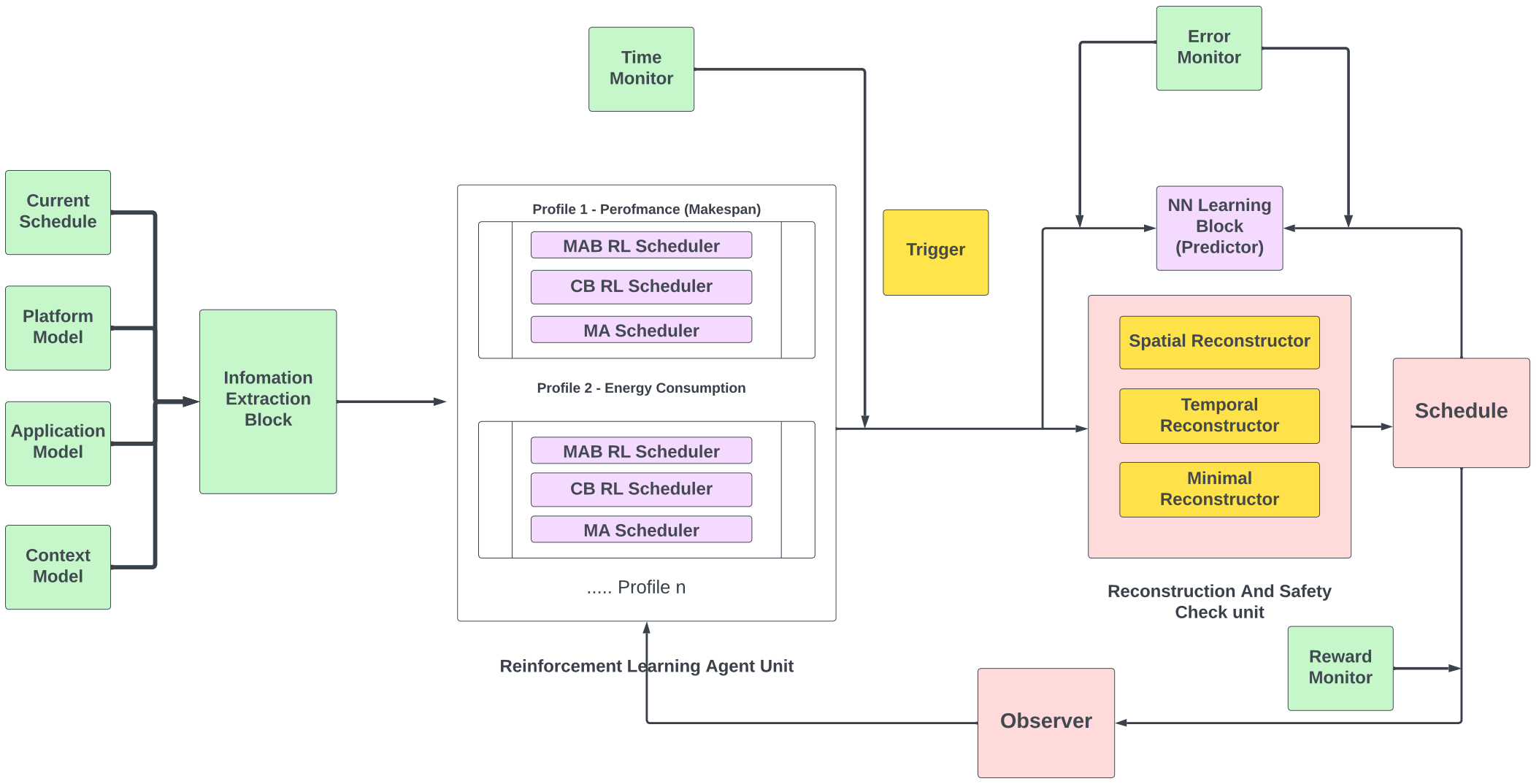

Figure 13: Evaluation of online learning unit component

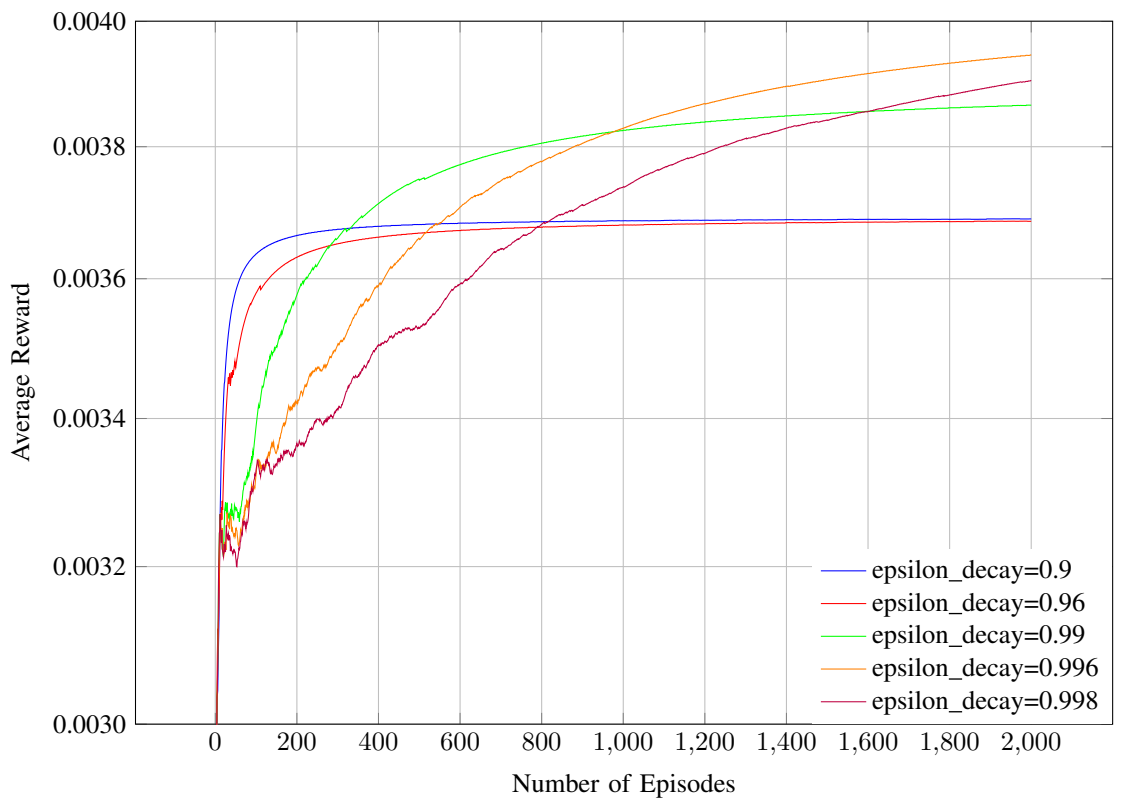

Figure 14: Epsilon decay selection

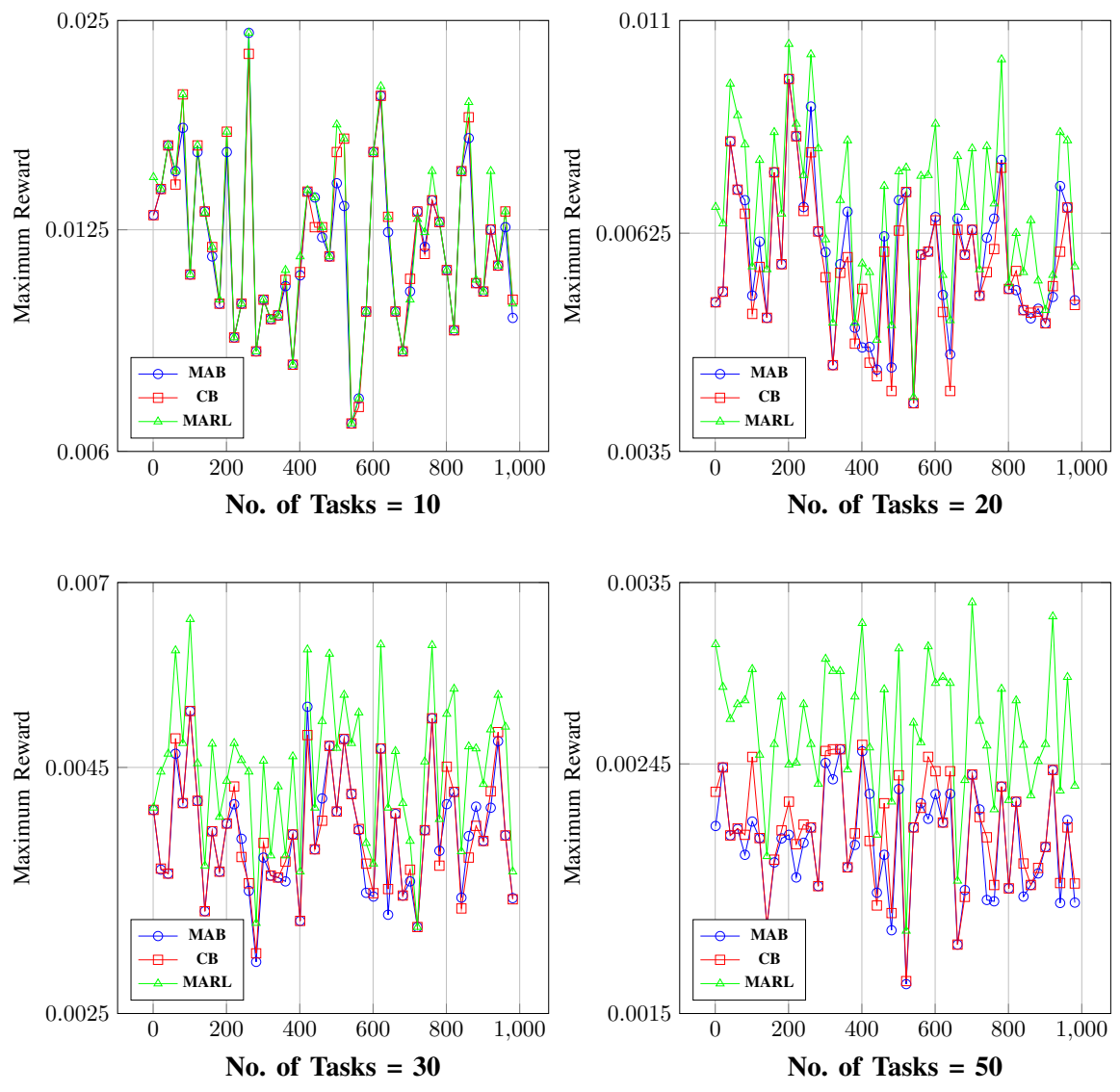

Figure 16: Maximum reward achieved per context for different task numbers

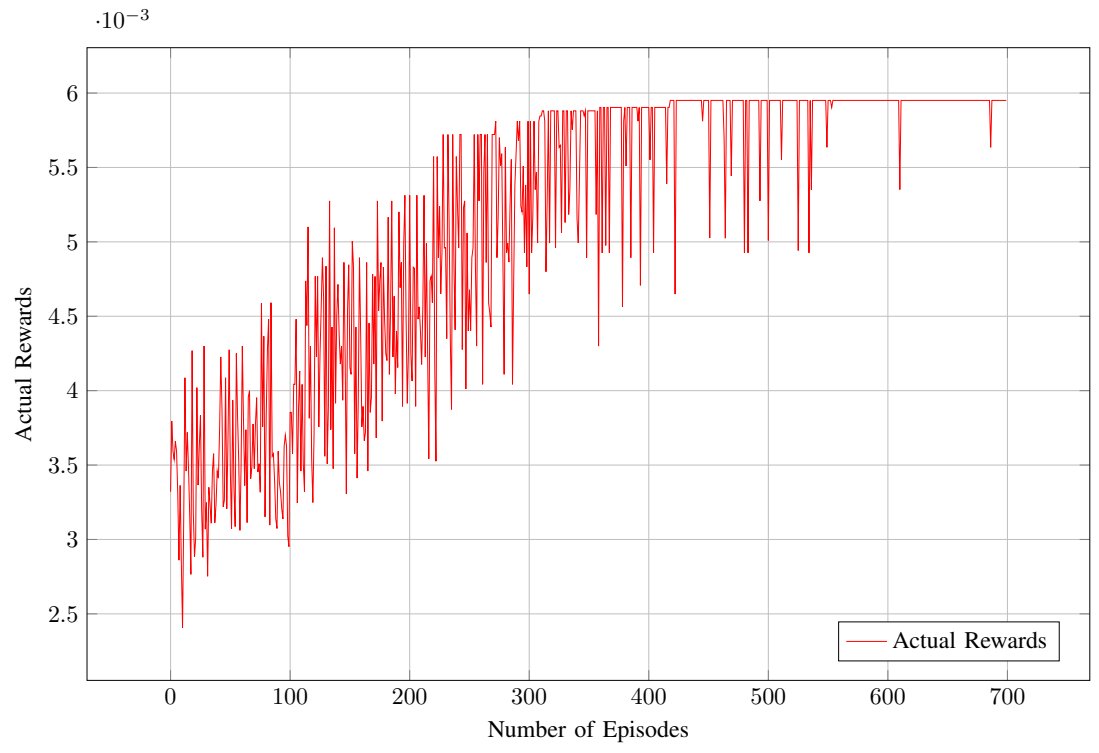

Figure 15: Exploitation and exploration for MARL

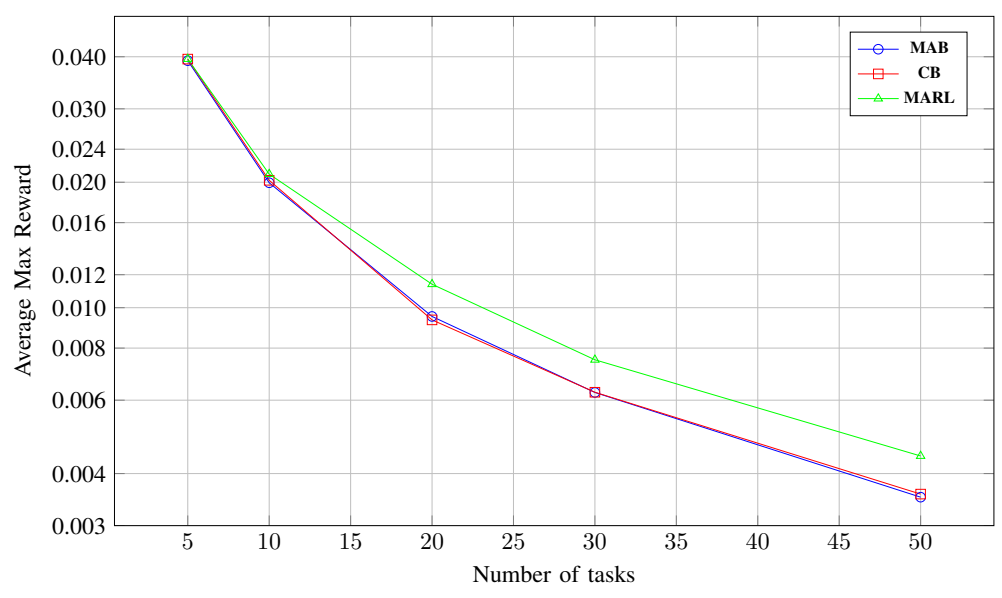

Figure 17: Comparison of average max rewards

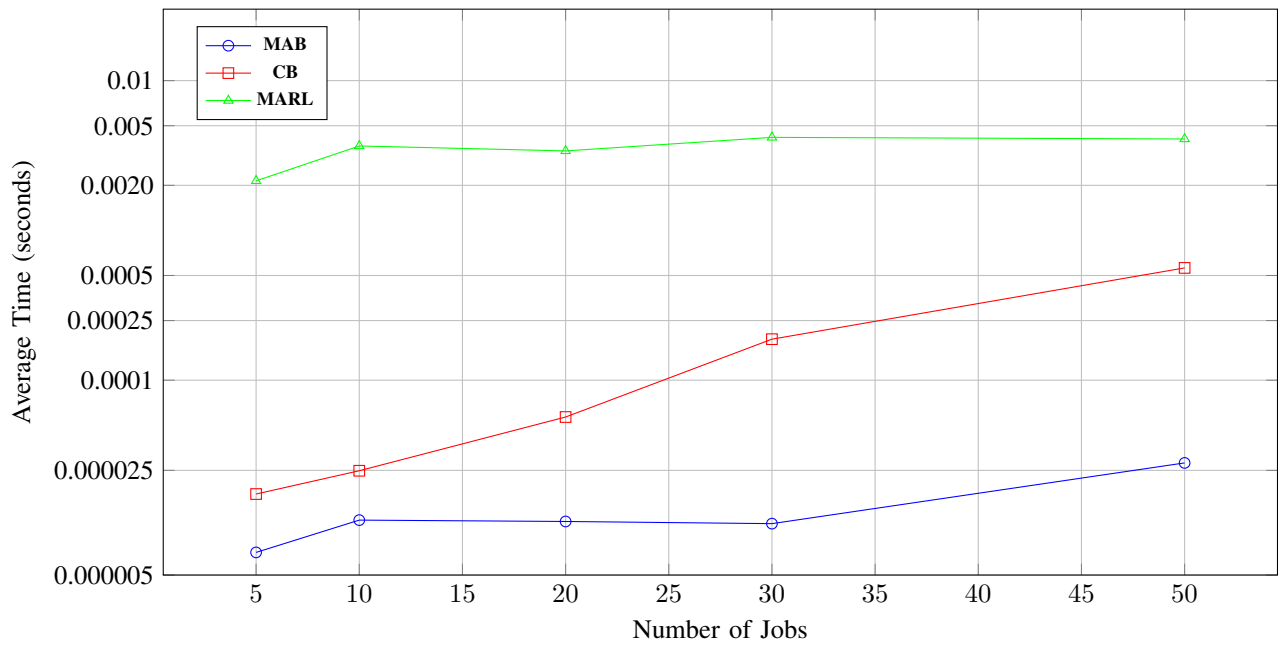

Figure 18: Comparison of decision making with time

Fig.19 show the prediction error with time of the NN attached to the online learning unit with the CB and MARL models respectively. The results show that the predictor is learning and reducing error values with every episode which confirms that it is learning with time.

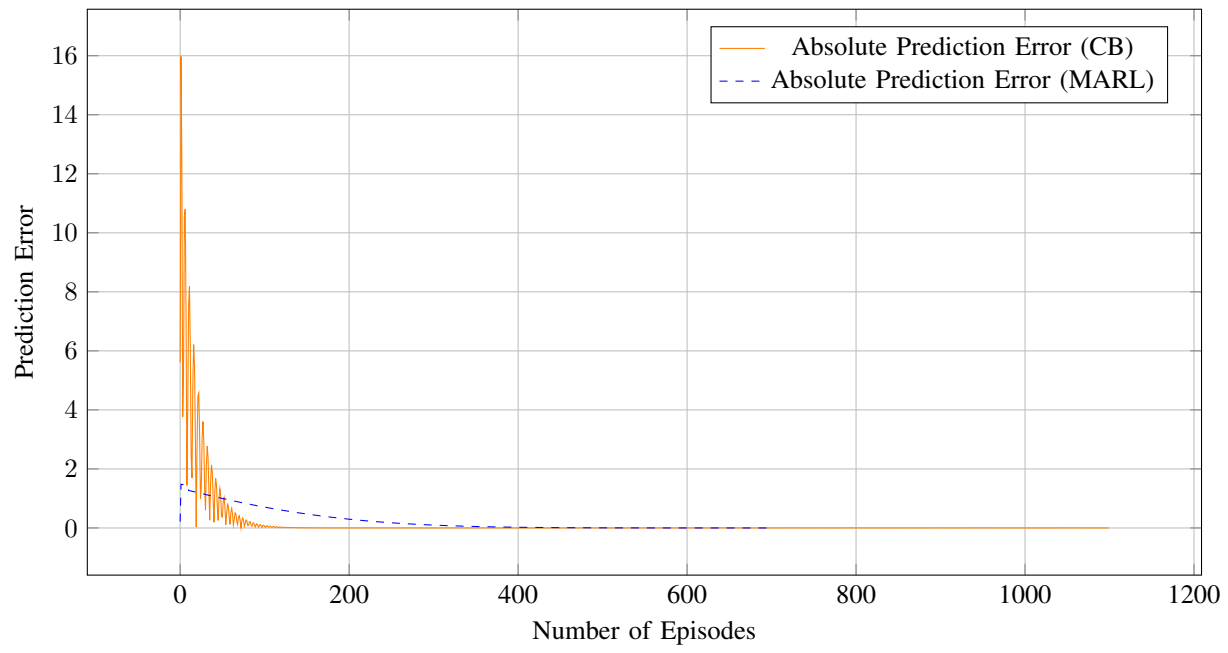

Figure 19: Absolute prediction error for CB and MARL

The online learning unit was employed to retrain the spatial priorities AI inference to enhance its decision-making process. The evaluation was conducted across different task configurations. The performance was assessed based on the makespan of the generated solutions both before and after training as the same 1,000 samples that were used to evaluate the performance of the AI spatial inference taken from the dataset were used to evaluate the performance difference. Fig. 20 presents a comparative analysis of the makespan before and after the online learning retraining process. While the y-axis denotes the makespan values. The results indicate that the inference model does not degrade after training, as the makespan remains the same or improves in most cases. This finding is critical, as it confirms that online learning does not negatively impact model performance, ensuring that retraining can be safely applied for adaptation without compromising solution quality.

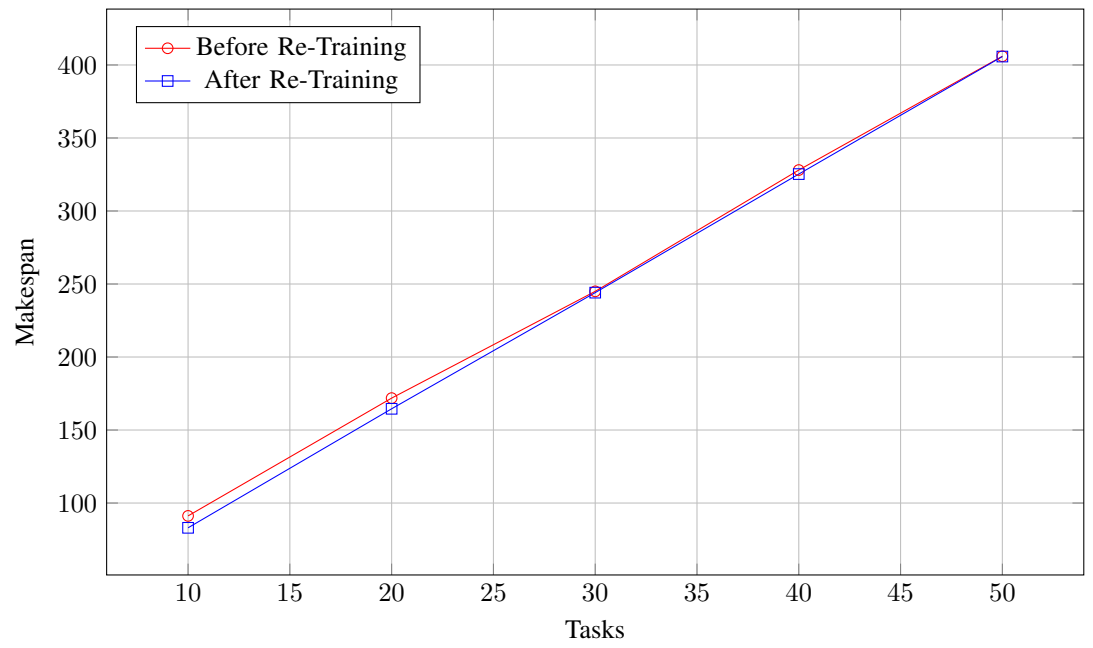

Figure 20: Comparison of makespan before and after online learning retraining for different task configurations.

Fig.21 presents a comparative analysis of the makespan results across different AI inference stages, MARL, and task deadlines. The primary objective of this study was to assess the impact of online retraining on AI spatial inferences in scheduling problems, ensuring that retrained inferences perform at least as well as or better than their pre-trained counterparts.

In this experiment, a total of 50,000 scheduling problems were generated, with 10,000 scheduling problems for each unique number of tasks. Each set of 10,000 problems was derived by selecting 1,000 original random schedules and injecting 10 context slack events into each, creating diverse scheduling variations. The experiment aimed to test whether the AI inference system could adapt and improve its scheduling efficiency through an online retraining process.

The pre-training performance of the AI inferences is depicted in the figure as the red curve, which consistently fails to meet the required deadlines (dashed black line). All 10,000 scheduling scenarios per task set exceeded their respective deadlines, triggering the necessity for an online learning intervention. To address this, the MARL search algorithm (blue curve) was employed to generate optimized scheduling solutions, which were subsequently used to retrain the AI inference models.

To facilitate effective retraining, 9,000 of the scheduling samples were allocated for training the online learning unit, ensuring that the model learned from its failures. The remaining 1,000 schedules were reserved for validation, following a systematic selection approach. Specifically, each original schedule generated 10 variations, out of which 9 were used for training, and 1 was retained as a test case. This methodology ensured that the AI was evaluated on unseen scheduling variations, rigorously assessing its generalisation capability beyond the training set.

After retraining, the AI inference performance, shown in green, demonstrates significant improvement. Unlike the pre-training results, the AI inferences after retraining achieve makespans that closely align with the MARL results and successfully meet the deadlines across all task sizes. This confirms that the online retraining mechanism effectively enhances scheduling efficiency without degrading performance.

The experimental setup and evaluation method reflect real-world scheduling scenarios, where failing schedules require corrective learning. However, this study takes a more rigorous approach by ensuring that test schedules are never part of the training set, providing a more robust validation of the online learning system's capability.

Fig.21 demonstrates the success of the online retraining mechanism in reducing the makespan of AI-generated schedules, validating the effectiveness of the proposed learning framework.

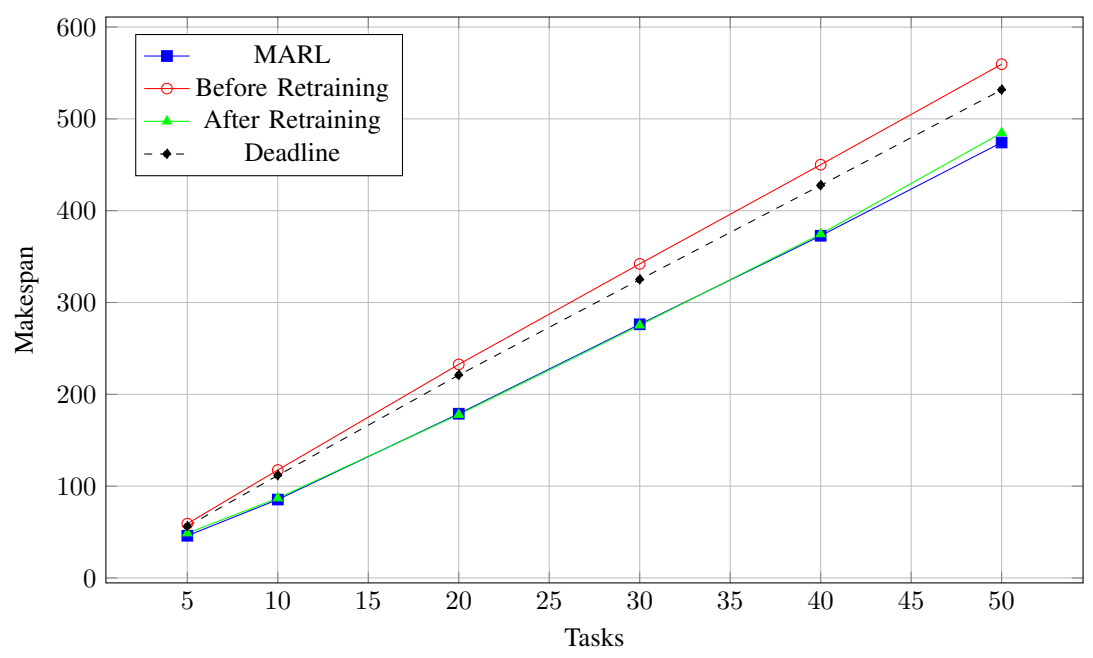

Figure 21: Comparison of makespan across different AI inference stages, MARL

## VII. Conclusion and Future Work

This study has introduced a novel approach for enhancing metascheduling applications by integrating an adaptive online learning unit, designed to improve scheduling efficiency and decision-making processes. As part of a broader research initiative focused on the application of machine learning techniques in metascheduling, this work specifically addressed the limitations of offline training by leveraging RL in an online setting. The experimental results demonstrate that the proposed online learning unit effectively enhances scheduling performance by dynamically refining AI inferences in response to real-time variations.

The comparative analysis of RL models revealed that while all models performed similarly in low-complexity scheduling tasks, their effectiveness diverged as task complexity increased. The MARL model exhibited superior reward optimization but incurred higher computational costs, highlighting the trade-offs between solution quality and execution efficiency. Furthermore, the integration of an AI-based prediction unit demonstrated a progressive reduction in prediction error over time, validating its learning capabilities.

Crucially, the study confirmed that the online learning unit could retrain AI spatial inferences without performance degradation. The results indicate that the retrained models achieved makespans comparable to those generated by MARL while successfully meeting strict scheduling deadlines. This underscores the adaptability and robustness of the proposed framework, ensuring reliable scheduling performance in dynamic and safety-critical environments.

For future work, the chosen RL search algorithm configuration, utilizing an epsilon-greedy mechanism, allows for independence between search epochs. This independence enables full parallelization of the algorithm, opening avenues for more efficient execution. Further exploration will focus on deploying the system on cloud-based or distributed architectures with higher clock rates than current local implementations. This shift would facilitate real-time execution of the online learning model, enabling immediate fault recovery and significantly enhancing system responsiveness in dynamic scheduling environments.

# VIII. Biography Section

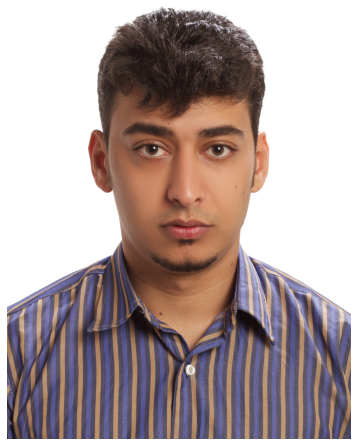

**Samer Alshaer** is currently a PhD candidate in Computer Engineering at the University of Siegen, Germany (since 2021), where he also works as a Research Assistant in the Embedded Systems division. He is a recipient of the prestigious DAAD research scholarship and winner of the QRCE Industry-Academia Linkage Competition. Prior to his doctoral studies, Samer served as a Research Assistant at the German Jordanian University (GJU). He earned his M.Sc. in Computer Engineering from GJU in 2019 and graduated first of his class with a B.Sc. in Mechatronics Engineering from Philadelphia University in 2014. His research focuses on intelligent scheduling systems, machine learning, reinforcement learning, and time-triggered architectures. He has authored multiple peer-reviewed publications in these fields.

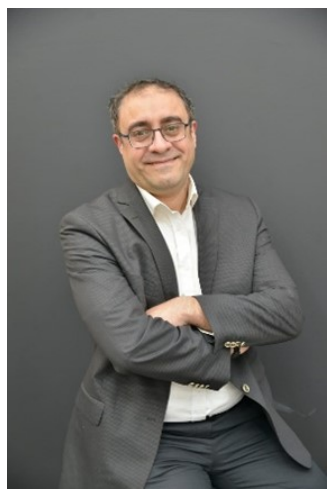

**Dr. Ala' Khalifeh** currently holds the position of President's Advisor for University Campus Life and diversity at the German Jordanian University (GJU), where he also works as a Professor of Electrical Engineering . Dr. Khalifeh received the prestigious Fulbright Scholarship in 2005, which enabled him to pursue his doctoral degree from the University of California-Irvine in the United States of America. In September 2021, Dr. Khalifeh became a Fellow of the Innovation Leaders Fellowship (LIF) Program run by the Royal Academy of Engineering - UK. Dr. Khalifeh's research focuses on emerging technologies including the Internet of Things, artificial intelligence, datal analytics, cloud computing and wireless sensor networks. Recognizing Dr. Khalifeh's research impact, he recently was recognized by Stanford-Elsevier Ranking of Top 2% Top-cited Researchers Includes in 2024.

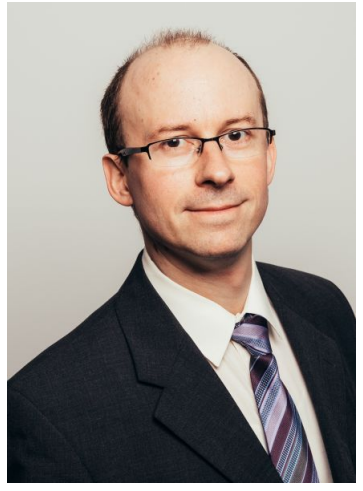

**Prof. Dr.-Ing. Roman Obermaisser** received the master's and Ph.D. degrees in computer sciences from the Vienna University of Technology, in 2001 and 2004, respectively, under the supervision of Prof. Hermann Kopetz. He was a Research Advisor with the Vienna University of Technology. He is currently a Full Professor with the Division for Embedded Systems, University of Siegen. He wrote a book on an integrated time-triggered architecture published by Springer-Verlag, USA. He is the author of several journal articles and conference publications. He has also participated in numerous EU research projects, such as SAFEPOWER, universAAL, DECOS, and NextTTA. He was the Coordinator of European research projects, such as DREAMS, GENESYS, and ACROSS. In 2009, he received the Habilitation ("Venia Docendi") Certificate for technical computer science.